\documentclass{article}
\usepackage{times}
\usepackage{hyperref}
\usepackage{url}
\usepackage{comment}
\usepackage{graphicx}
\usepackage{enumitem}
\usepackage{natbib}
\usepackage{geometry}

\usepackage[capitalize]{cleveref}  

\crefname{nlem}{Lemma}{Lemmas}
\crefname{nprop}{Proposition}{Propositions}
\crefname{ncor}{Corollary}{Corollaries}
\crefname{nthm}{Theorem}{Theorems}
\crefname{assumption}{Assumption}{Assumptions}

\RequirePackage[usenames,dvipsnames]{color}

\usepackage{jhh-misc2}

\usepackage[accepted]{icml2015}

\boldshortcuts

\allowdisplaybreaks[3]

\icmltitlerunning{Risk and Regret of Hierarchical Bayesian Learners}

\begin{document}

\twocolumn[
\icmltitle{Risk and Regret of Hierarchical Bayesian Learners}

\icmlauthor{Jonathan H.~Huggins}{jhuggins@mit.edu}
\icmladdress{Computer Science and Artificial Intelligence Laboratory, MIT}

\icmlauthor{Joshua B.~Tenenbaum}{jbt@mit.edu}
\icmladdress{Brain and Cognitive Science Department, MIT}

\icmlkeywords{learning theory, online learning, risk bounds, regret bounds, Bayesian hierarchical models}

\vskip 0.3in
]

\begin{abstract}
Common statistical practice has shown that the full power of Bayesian methods 
is not realized until hierarchical priors are used, as these
allow for greater ``robustness'' and the ability to 
``share statistical strength.'' 
Yet it is an ongoing challenge to provide a learning-theoretically 
sound formalism of such notions that:
offers practical guidance concerning when and how
best to utilize hierarchical models; 
provides insights into what makes for a good hierarchical prior;
and, when the form of the prior has been chosen, can guide the
choice of hyperparameter settings.
We present a set of analytical tools for understanding
hierarchical priors in both the online and batch learning settings. 
We provide regret bounds under log-loss, which
show how certain hierarchical models compare, in retrospect,
to the best single model in the model class.
We also show how to convert a Bayesian log-loss regret bound
into a Bayesian risk bound for any bounded loss, a result which
may be of independent interest.
Risk and regret bounds for Student's $t$ and hierarchical Gaussian
priors allow us to formalize the concepts of ``robustness'' 
and ``sharing statistical strength.''
Priors for feature selection are investigated as well. 
Our results suggest that the learning-theoretic benefits of 
using hierarchical priors can often come at little cost 
on practical problems. 
\end{abstract}

\section{Introduction}
\label{sec:introduction}

There are two standard justifications for the use of hierarchical 
models. 
The first is that they allow for the representation of greater uncertainty 
by placing ``hyperpriors'' on the hyperparameters of the prior 
distribution~\citep{Berger:1985,BS:2000,Gelman:2013}.
By explicitly modeling the additional uncertainty, there is greater
``robustness'' to misspecification and unexpected data. 
The second is that hierarchical models permit the 
``sharing of statistical strength'' 
between related observations or cohorts~\citep{Gelman:2013}.
For example, take the recent ``Big Bayes Stories'' special issue 
of the journal \emph{Statistical Science},
which was comprised of short articles describing successful applications 
of Bayesian models to a diverse range of problems, including 
political science, astronomy, and public health~\citep{Mengersen:2014}.
Most of the Bayesian models were hierarchical, and the need for
robustness and sharing of statistical strength because of limited 
data were commonly cited reasons by the practitioners for choosing a
hierarchical Bayesian approach. 
\citet{Gelman:2006} and \citet{Gelman:2013} both contain further examples of
problems in which hierarchical modeling is critical to obtaining
high-quality inferences. 

\begin{figure}[tbp]
\begin{center}
\includegraphics[width=.225\textwidth]{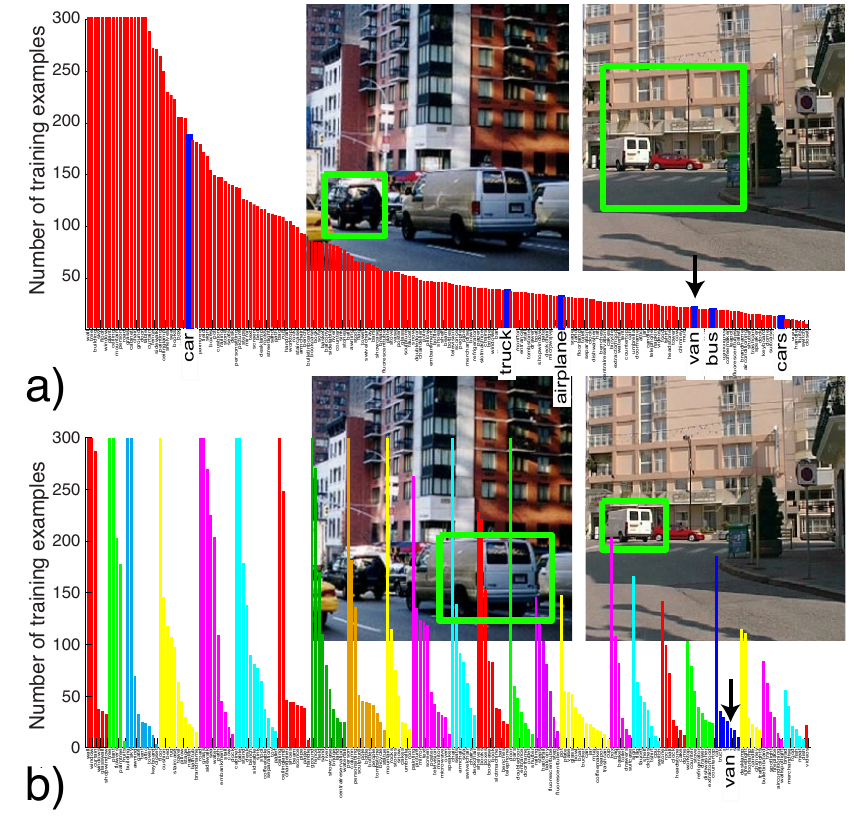}
\includegraphics[trim = 0mm 0mm 0mm 0mm, width=.225\textwidth]{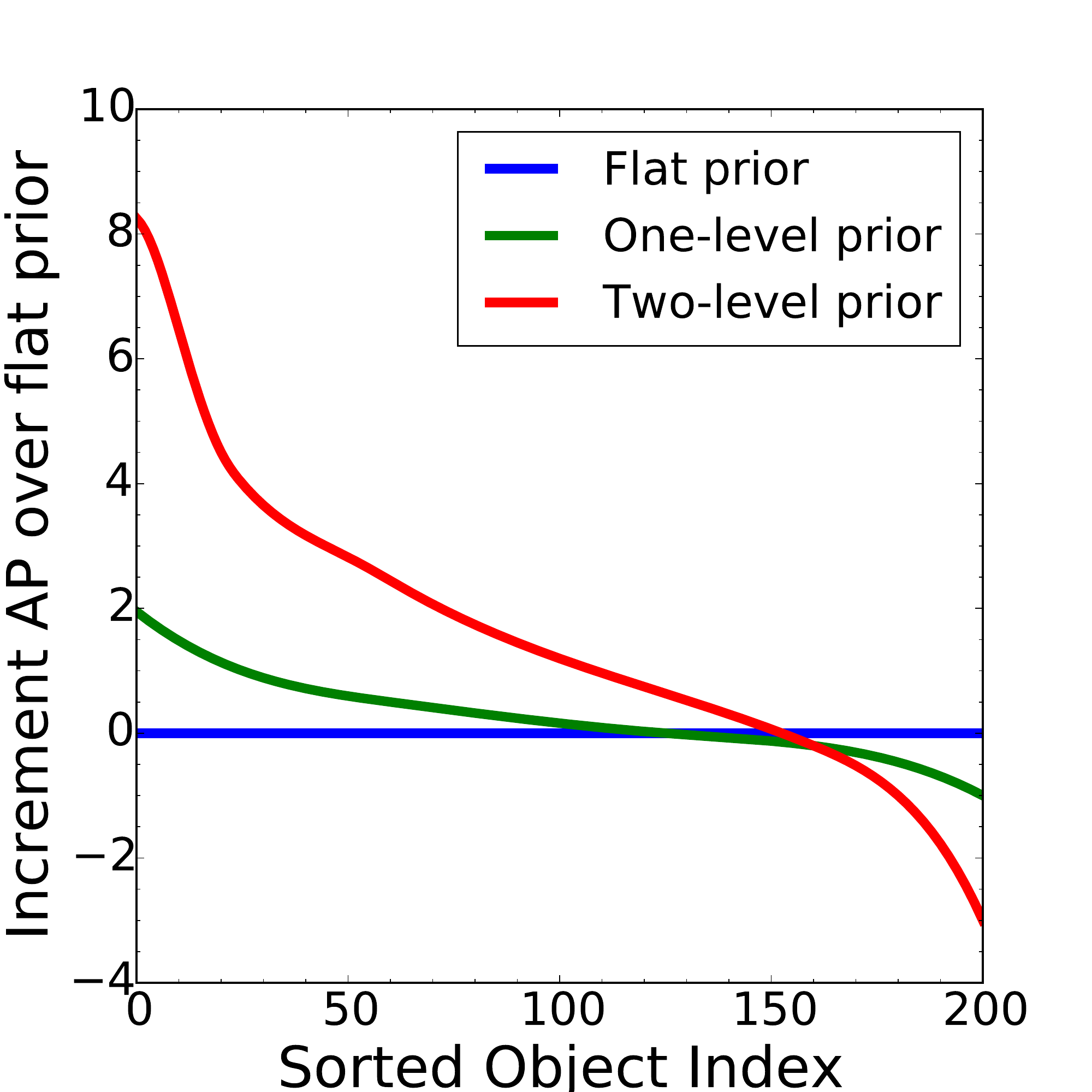}
\caption{\textbf{Right:} \textbf{a)}~Distribution of training examples per object class. 
\textbf{b)}~Same as a), but with objects grouped by visual appearance. 
\textbf{Left:} Improvement in classification accuracy of hierarchical 
models compared to flat model. 
Object categories are sorted by improvement.
Reproduced and reconstructed from \citet{Salakhutdinov:2011}.}
\label{fig:cvpr}
\end{center}
\vspace{-6mm}
\end{figure}

Within the machine learning and vision literature, \citet{Salakhutdinov:2011} 
offers an illustrative case study in the benefits and the pitfalls of employing 
a hierarchical model. 
The motivation of \citet{Salakhutdinov:2011} was that, for 
image classification tasks, some categories of objects
(e.g., ``car'' or ``dog'')  have many labeled positive and negative
examples while other, visually related, categories (e.g., ``bus'' 
or ``anteater'') have only a few labeled examples. 
\cref{fig:cvpr}(right, a) shows the distribution of training
examples for the 200 object categories used while 
\cref{fig:cvpr}(right, b) shows the same distribution, but now
objects are grouped with those with similar appearances.
In both cases, the distributions are fat-tailed: there are a few categories
with many training examples and many categories with a few training examples. 
It was hypothesized that by using a hierarchical Bayesian model, 
the classes with large amounts of labeled data could be used to construct better
classifiers for the classes with small amounts of labeled data. 

The model used by \citet{Salakhutdinov:2011}, 
which we will analyze in \cref{sec:share-strength},
consisted of a hierarchical Gaussian prior with a logistic regression likelihood.
Two-level, one-level, and flat priors were all tested.
The purpose of using the two-level prior was that it was able to encode
information about which object classes had visually similar objects
(e.g., car and track, dog and horse).
\cref{fig:cvpr}(left) compares the predictive performance of
this two-level hierarchical prior with the two more impoverished priors.
Observe that the one-level and two-level priors both improve performance on 
most object classes compared to the flat prior, but not all. 
Furthermore, the two-level prior always leads to greater improvement than
the one-level prior on object classes where a hierarchical model helps, but
also almost always leads to a greater degradation in performance on
object classes where the hierarchical models decrease performance. 
Why the different performance characteristics for the two 
hierarchical models? 
Why do some categories have improved accuracy while others
decreased accuracy? 
In a post-hoc analysis, \citet{Salakhutdinov:2011} note that
the ``objects with the largest improvement...borrow visual appearance
from other frequent objects'' while ``objects with the largest
decrease [such as `umbrella' and `merchandise'] are 
abstract, and their visual appearance is very different from 
other object categories.''

The results just described lead to numerous theoretical
questions of practical consequence:
\begin{enumerate}[label=Q\arabic*]
\item Can we formalize why for some object classes there was a beneficial sharing 
of statistical strength, while for other classes the sharing was detrimental? \label{q:sharing}
\item Can we understand when a flat model should be preferred to a hierarchical one to
avoid unfavorable sharing? \label{q:flat}
\item More generally, can we obtain guidance on the best type of prior for the problem at hand? 
Perhaps a different hierarchical prior would have been better suited to 
learning the image classifiers. 
For example, could placing hyperpriors on the variance parameters lead
to greater ``robustness'' for object categories such as `umbrella' and `merchandise,' 
 whose  visual appearance differs from other object categories? \label{q:prior-guidance}
\item Once the form of the prior has been chosen, how should hyperparameters 
be set to maximize learning? 
The settings of the variance hyperparameters was left unspecified by 
\citet{Salakhutdinov:2011}, and it is not clear a priori how they should be set, 
or how much effect their choice will have on learning.  \label{q:hypers}
\end{enumerate}
While we have primarily framed these questions in terms of a single model from one paper,
this focus was simply for concreteness.
Similar results leading to the same types of questions can be found in
the numerous articles that make use of hierarchical Bayesian methods.
For example, one might instead consider the hierarchical models have been 
used in political science for analyzing polling and census data to predict 
election outcomes \citep{Ghitza:2013} and
in demography for predicting population growth, life expectancy, and 
fertility rates \citep{Raftery:2013,Raftery:2012,Alkema:2011}. 

In this paper we seek to answer the questions just posed
in terms of two learning-theoretic quantities: regret (in online learning) and 
statistical risk (in batch learning).
The online learning setting applies, for example, to the demography
applications and election prediction while the batch setting is 
relevant to the image classification problem
as well as election prediction (whether the online or batch analysis 
applies to election prediction depends on how the problem is formulated). 

In the online learning 
framework~\citep{Dawid:1999,CesaBianchi:2006}, no assumptions are 
made about the data-generating mechanism.
Inputs are presented to the learner one by one.
After receiving each input the learner predicts the output, 
then suffers a loss after observing the true output. 
The goal of the learner is to not do
much worse (i.e., have large regret) compared to 
a fixed class of predictors. 
Online learning guarantees are attractive for the analysis
of hierarchical Bayesian models because such models are 
so often used in exactly those circumstances when orthodox
Bayesian justifications do not apply: typically the
modeler does not think that her model reflects 
the true data generating process, but is instead employing 
hierarchical methods either to increase robustness against 
a poor choice of hyperparameters or to speed learning
by allowing for the sharing of statistical strength between
populations. 

Regret bounds, however, do not themselves give any generalization guarantees
about how the learner will perform on future data.
Statistical risk bounds provide guarantees
about the learner's expected loss on unseen examples
by making assumptions about how the
data are generated --- for example, from an \iid\ or 
strongly mixing process. 
Although there is a stochastic assumption, risk bounds
also do not assume that the data is generated according
to the model used.
We derive a general result for transferring Bayesian regret bounds
to risk bounds for bounded losses.

Regret bound for a number of Bayesian models have
previously been developed \citep{Vovk:2001,Kakade:2004,Kakade:2005,
Banerjee:2006,Banerjee:2007,Seeger:2008},
with a particular focus on regression and simple priors
such as independent Gaussian distributions for each regression coefficient. 
For a discussion of more general (but asymptotic) Bayesian regret
bounds for exponential families and other sufficiently
``regular'' model classes, see \citet[Chapter 8]{Grunwald:2007}.
We follow the approach originally taken in \citet{Kakade:2004},
and further explored in~\citet{Kakade:2005}, \citet{Banerjee:2006},
and \citet{Seeger:2008}, which applies to a large class of 
Bayesian generalized linear models (GLMs).
We extend the technique to apply to certain non-GLM likelihoods as well,
including to multi-class logistic regression.
All proofs are deferred to the Appendices.

We answer Questions \ref{q:sharing}-\ref{q:hypers} in some important cases
by deriving regret bounds for three types of hierarchical priors. 
First, we consider the use of an inverse gamma hyperprior
for the Gaussian prior's variance parameter and 
demonstrate that the hyperprior leads to greater 
robustness to data that is well-explained by setting
the GLM parameter vector to have very large $\ell^{2}$
norm.
Next, we analyze hierarchical Gaussian models
that allow for the sharing of statistical 
strength.
Our results, which complement existing work on
transfer and multitask learning theory
\citep{Baxter:1997,BenDavid:2003,Pentina:2014}, 
show that when the parameters with
small regret for a collection of related tasks are 
either (a) similar or (b) not unexpected under the 
prior, then the hierarchical model has a smaller
regret bound than assuming the tasks are independent. 
Finally, we show that spike-and-slab priors can
exploit sparse parameters with small regret.

\section{Bayesian Online Learning}

In online learning, the learner must predict (a distribution over)
$y \in \mcY \subseteq \reals$ 
after observing 
$\bx \in \mcX \subseteq \reals^{n}$. 
In this paper, we assume the prediction is made according 
to a generalized linear model (GLM)
$p(y \given \bx, \btheta) = p(y \given \btheta \cdot \bx)$, 
where $\btheta \in \Theta \subseteq \reals^{n}$ is a parameter vector to be chosen.
GLMs provide significant modeling flexibility, and
the class of GLM models and priors we analyze include a range of models 
used in real-world scientific applications \citep{Gelman:2006,Gelman:2013}. 
Two widely used GLMs are the logistic regression likelihood
\[
p(y \given \btheta, \bx) &= \frac{1}{1 + \exp(y\btheta\cdot \bx)}, & y \in \{-1,1\},
\label{eq:logistic-regression}
\]
and the Gaussian linear regression likelihood 
$p(y \given \btheta, \bx) = \distNorm(y \given \btheta \cdot \bx, \sigma^{2})$, $y \in \reals$.
Since we are taking a Bayesian approach, 
we place a prior density $p_{0}(\btheta)$ on $\Theta$,
with corresponding distribution $P_{0}$.\footnote{Throughout, we use lowercase letters for densities and uppercase letters to denote the corresponding measures.}
At time step $t$, the learner observes $\bx_{t}$, outputs a 
distribution over $\mcY$, then observes $y_{t} \in \mcY$.
The Bayesian (model average) learner predicts $p(y \given \bx_{t}, Z_{t-1})$,
where $Z_{t} \defined \theset{(\bx_{1}, y_{1}), \dots, (\bx_{t}, y_{t})}$,
and then suffers the log-loss $-\ln p(y_{t} \given \bx_{t}, Z_{t-1})$.
Hence, the cumulative loss incurred is
\(
L_{Bayes}(Z_{T}) \defined \textstyle\sum_{t=1}^{T} -\ln p(y_{t} \given \bx_{t}, Z_{t-1}).
\)
If $Q$ is a distribution over $\btheta$, then
using $Q$ for prediction leads to loss on example $t$ of
$\ell_{t}(Q) \defined \EE_{Q}[-\ln p(y_{t} \given \bx_{t}, \btheta)]$
and hence cumulative loss
\(
L_{Q}(Z_{T}) \defined \EE_{Q}\left[\textstyle\sum_{t=1}^{T}-\ln p(y_{t} \given \bx_{t}, \btheta)\right]. 
\)
If $Q = \delta_{\btheta}$, then we write $L_{\btheta}$ instead of 
$L_{Q}$, so $L_{Q}(Z_{T}) = \EE_{Q}[L_{\btheta}(Z_{T})]$.
Our objective is to derive regret bounds of the form 
\[
\mcR(Z_{T}, \btheta) \defined L_{Bayes}(Z_{T}) -  L_{\btheta}(Z_{T}) \le B(\btheta) + C(T), \label{eq:generic-regret-bound}
\]
where $\mcR(Z_{T}, \btheta)$ is the \emph{regret} and $B(\btheta) + C(T)$ 
is a \emph{regret bound} depending on the choice of prior $P_{0}$. 
We aim for $C(T) = o(T)$, so that for a fixed $\btheta$,
the average loss $T^{-1}L_{Bayes}(Z_{T})$ is 
bounded by $T^{-1}L_{\btheta}(Z_{T}) + o(1)$. 

Our approach to bounding $L_{Bayes}(Z_{T})$ 
follows that of previous work on Bayesian GLM regret bounds 
with log-loss~\citep{Kakade:2004,Kakade:2005,Seeger:2008}, 
relying on the following well-known result:
\bnprop[\citet{Kakade:2004,Banerjee:2006}] \label{prop:compression}
The Bayesian cumulative loss is bounded as
\[
L_{Bayes}(Z_{T}) \le L_{Q}(Z_{T}) + \kl{Q}{P_{0}}. \label{eq:compression}
\]
\enprop
For the GLM model $p(y \given \btheta \cdot \bx)$, define 
$f_{y}(z) \defined - \ln p(y \given \btheta \cdot \bx = z)$. 
We make two assumptions throughout the remainder of the paper
 (they will usually not be stated explicitly):
\[
|f_{y}''(z)| &\le c \quad  \text{for all $y, z$} \tag{A1} \label{eq:a1} \\
\|\bx_{t}\|_{2} &\le 1 \quad \text{for all $t$}. \tag{A2} \label{eq:a2}
\]
The first assumption can be understood as requiring the
likelihood to be sufficiently smooth.
The second assumption sets the scale of the problem, which
is necessary since scaling $\bx_{t}$ up requires scaling $\btheta$
down, and vice-versa: $p(y \given C^{-1}\btheta \cdot C\bx) = p(y \given \btheta \cdot \bx)$
for any $C \ne 0$. 
Note that for the Gaussian linear regression model with 
variance $\sigma^{2}$ and the logistic regression model, 
\eqref{eq:a1} holds with $c = 1/\sigma^{2}$ and $c = 1/2$,
respectively. 
Proposition \ref{prop:compression} leads to the following theorem for 
obtaining regret bounds for the Bayesian model average learner.
\bnthm[Bayesian regret meta-theorem] \label{thm:meta-theorem}
Let $Q_{\btheta^{*},\bphi}$ be a distribution with parameter
$\bphi \in \Phi \subseteq \reals^{d}$ (written $\phi$ if $d = 1$)
and mean $\btheta^{*}$. 
If \eqref{eq:a1} and \eqref{eq:a2} hold, then for all $\bphi$,
\(
\mcR(Z, \btheta^{*}) \le \frac{Tc}{2}\|\var_{Q_{\btheta^{*},\bphi}}[\btheta]\| + \kl{Q_{\btheta^{*},\bphi}}{P_{0}},
\)
where $\|\cdot\|$ is the spectral norm. 
In particular, if the components of $\btheta^{*}$ are uncorrelated,
then $\|\var_{Q_{\btheta^{*},\bphi}}[\btheta]\| = \sup_{i} \var_{Q_{\btheta^{*},\phi}}[\theta_{i}]$
\enthm
\cref{thm:meta-theorem} is our first main result and will be repeatedly applied in \cref{sec:applications} 
by choosing an appropriate $Q_{\btheta^{*},\bphi}$ and then optimizing $\bphi$.
Although the bound appears to be linear in $T$, typically $\bphi$ can be
chosen such that $\|\var_{Q_{\btheta^{*},\bphi}}[\btheta]\| = \Theta(T^{-1})$ 
and $\kl{Q_{\btheta^{*},\bphi}}{P_{0}} = \Theta(\ln T)$, leading to a logarithmic
regret bound.
\cref{thm:meta-theorem} provides an  attractive approach to deriving regret 
bounds because there is no need to work directly with the posterior,
which is often analytically intractable. 
For example, there is no closed-form expression for the posterior 
of the Bayesian logistic regression model.
The theorem generalizes the approach originally taken in 
\citet{Kakade:2004}, in which a Gaussian prior for $\btheta$ was considered:
\bnthm[Gaussian regret~\citep{Kakade:2004}] \label{thm:gaussian-regret}
If $\btheta \dist \distNorm(0, \sigma^{2}I)$, then $\mcR(Z,\btheta^{*})$ is
bounded by
\(
R_{Bayes}^{G}(Z, \btheta^{*})
&\defined  
   \frac{1}{2\sigma^{2}}\|\btheta^{*}\|^{2} 
   + \frac{n}{2} \ln\left(1 + \frac{Tc\sigma^{2}}{n}\right).
\)
\enthm

\subsection{Beyond GLMs}

\cref{thm:meta-theorem} follows from a more general result, \cref{thm:super-meta-theorem},
which allows for non-GLM likelihoods. 
Specifically, instead of the likelihood being a GLM, we assume 
the likelihood can be written in the form $p(y \given \bx, \bxi, \bpsi) = p(y \given \bxi\bx, \bpsi)$,
where $\bxi \in \reals^{n' \times n}$ is a matrix and $\bpsi \in \reals^{n''}$. 
The full parameter vector is $\btheta = (\bxi, \bpsi) \in \reals^{N}$, $N \defined nn' + n''$
(implicitly flattening the matrix $\bxi$). 
Let $f_{y}(\bz) \defined -\ln p(y \given (\bxi \bx, \bpsi) = \bz)$. 
We require the following assumption in place of \eqref{eq:a1}:
\[
\|f_{y}''(\bz)\| &\le c \quad \text{for all $y,\bz$} \tag{A1'} \label{eq:a1'},
\]
where $f_{y}''(\bz)$ denotes the matrix of second partial derivatives (Hessian).

\bnthm[Generalized Bayesian regret meta-theorem] \label{thm:super-meta-theorem}
Let $Q_{\btheta^{*},\bphi}$ be a distribution with parameter
$\bphi \in \Phi \subseteq \reals^{d}$ and mean $\btheta^{*}$. 
If \eqref{eq:a1'} and \eqref{eq:a2} hold, then for all $\bphi$,
\(
\mcR(Z, \btheta) \le \frac{Tc(n' + n'')}{2}\|\var_{Q_{\btheta^{*},\bphi}}[\btheta]\| 
 + \kl{Q_{\btheta^{*},\bphi}}{P_{0}} ,
\)
\enthm

Of particular interest is that  \cref{thm:super-meta-theorem} can handle
multi-class logistic regression (MLR).
In multi-class regression, each example $\bx_{t}$ has one of $K$ labels
$y_{t} \in \{1,\dots,K\}$, indicating which class the example belongs to.
For MLR, each class $k$ has an associated parameter $\btheta^{(k)}$.
The parameters are combined into a single likelihood:
\[
p(y_{t} \given \btheta, \bx_{t}) 
= \frac{\exp(\btheta^{(y_{t})}\cdot \bx)}{\sum_{k=1}^{K}\exp(\btheta^{(k)}\cdot \bx_{t})}.
\]
\bnthm[MLR Gaussian regret] \label{thm:mlr-gaussian-regret}
If $\btheta^{(k)} \dist \distNorm(0, \sigma^{2}I)$, $k=1,\dots,K$, 
then using the MLR likelihood guarantees that $\mcR(Z, \btheta^{*})$ is
bounded by
\(
R_{Bayes}^{mlr-G}(\btheta^{*}, Z)
&\defined  
   \frac{1}{2\sigma^{2}}\|\btheta^{*}\|^{2} 
   + \frac{nK}{2} \ln\left(1 + \frac{TKc\sigma^{2}}{n}\right).
\)
\enthm

\section{Risk Bounds}
\label{sec:risk}

While online regret bounds are attractive because they make no assumptions
about the data-generating process, it is also desirable to have 
risk bounds in the batch setting since risk bounds provide generalization
guarantees for unseen data. 
We now develop a connection between regret and risk bounds via a 
PAC-Bayesian analysis~\citep{McAllester:2003a,
Audibert:2007,Catoni:2007}.
Such bounds also have the benefit of applying to any bounded
loss (e.g., the 0-1 loss for binary classification), which 
may be more task-relevant than the log-loss. 
In the batch setting, the data $Z_{T}$ are received all at 
once by the learner and are assumed to be distributed \iid\ 
according to some distribution $\mcD$ over $\mcX \times \mcY$:
$(\bx_{t}, y_{t}) \distiid \mcD$, $t=1,\dots,T$.
Let $\ell$ be a bounded loss function taking
a probability distribution over $\mcY$ and an element of
$\mcY$ as arguments. 
Without loss of generality assume $\ell \in [0,1]$. 
Writing $\ell_{\btheta}(\bx,y) \defined \ell(p(\cdot\given \bx, \btheta), y)$,
for any distribution $Q$ over $\Theta$, let 
\(
\mcL(Q) &\defined \EE_{(\bx,y)\dist \mcD}\EE_{\btheta \dist Q}[\ell_{\btheta}(\bx,y)]  \\
\hat\mcL(Q, Z_{T}) &\defined T^{-1}\textstyle\sum_{t=1}^{T} \EE_{\btheta \dist Q}[\ell_{\btheta}(\bx_{t},y_{t})]
\)
be, respectively, the expected and empirical losses under $Q$. 
PAC-Bayesian analyses consider the risk of the 
Gibbs predictor for the distribution $Q$ 
(i.e., sample $\btheta \dist Q$, predict with $p(\cdot\given \bx,\btheta)$),
not the model average over $Q$ (i.e., predict with 
$\int p(\cdot\given \bx,\btheta)Q(\dee\btheta)$). 
A typical  bound (specialized to the Bayesian setting)
is the following (here $p_{T}(\btheta) \defined p(\btheta \given Z_{T})$):
\bnthm[\citet{Audibert:2007}]  \label{thm:pac-bayes}
Fix $\kappa > \half$ and write $\kappa' \defined 2\kappa/(2\kappa - 1)$.
For any distribution $\mcD$, with probability at least $1 - \delta$ over 
samples $(\bx_{t},y_{t}) \distiid \mcD$,
\[
\begin{split}
\lefteqn{|\mcL(P_{T}) - \hat\mcL(P_{T}, Z_{T})|} \\
&\le T^{-1/2}\sqrt \kappa \sqrt{\kl{P_{T}}{P_{0}} + \ln \kappa'/\delta}. 
\end{split}
\]
\enthm
Combining the PAC-Bayesian risk bound with Bayesian regret bounds
leads to our second main result:
\bnthm \label{thm:pac-bayes-regret}
Assume that \eqref{eq:generic-regret-bound} holds
and fix $\kappa > \half$.
For any distribution $\mcD$, with probability at least $1 - \delta$ over 
samples $(\bx_{t},y_{t}) \distiid \mcD$, 
\[
\begin{split}
\lefteqn{|\mcL(P_{T}) - \hat\mcL(P_{T}, Z_{T})|} \\
&\le T^{-1/2} \sqrt \kappa \sqrt{B(\boldsymbol\htheta) + C(T) + \ln \kappa'/\delta},  \label{eq:regret-pac-bayes-bound} 
\end{split}
\]
where $\boldsymbol\htheta \defined \argmin_{\btheta} L_{\btheta}(Z_{T})$.
\enthm
An attractive feature of \cref{thm:pac-bayes-regret} is that the bound does 
not rely on understanding the posterior $P_{T}$, as is required by
a direct application of a PAC-Bayesian bound such as that given in
\cref{thm:pac-bayes}, which requires calculating $\kl{P_{T}}{P_{0}}$.
Yet the PAC-Bayesian regret bound remains data-dependent 
due to its dependence on the empirical risk minimizer (ERM) $p(y\given \bx,\boldsymbol\htheta)$. 

Examining the proof of \cref{thm:pac-bayes-regret}, it is easily seen that 
in fact any $\boldsymbol{\tilde\theta}$ such that 
$L_{\boldsymbol{\tilde\theta}}(Z_{T}) < L_{P_{T}}(Z_{T})$ can be
chosen in place of $\boldsymbol\htheta$.
Such alternative choices may lead to significantly tighter 
bounds and are particularly important, for example,
in the application of the theorem to the spike-and-slab prior
(cf.~\cref{sec:sparsity}), as the ERM parameter will in almost all circumstances 
satisfy $\|\boldsymbol\htheta\|_{0} = n$, which would lead to a 
poor generalization bound when $n$ is large. 

In words, \cref{thm:pac-bayes-regret} can be understood as stating that 
if the Bayesian (model average) learner has small log-loss regret compared
to the ERM,
then with high probability the Bayesian Gibbs predictor will generalize 
well if the loss function is bounded.
Or, as a slogan, the theorem shows that ``small regret in the online learning setting 
implies good generalization bounds in the batch setting.'' 
The theorem thus connects PAC-Bayesian bounds, Bayesian regret bounds,
and empirical risk minimization.

\section{Applications}
\label{sec:applications}

We now use \cref{thm:meta-theorem} to 
analyze hierarchical priors for robustness,
sharing of statistical strength, and feature selection.

\subsection{Hierarchical Priors for Robustness}
\label{sec:t-distribution}

In this section we answer questions \ref{q:flat}-\ref{q:hypers} as they
relate to hierarchical priors for robust inference, demonstrating
how, with a proper choice of hyperparameters, a hierarchical prior can
lead to increased robustness compared to a flat prior.
Specifically, we analyze a canonical use of a hierarchical prior --- 
to capture greater uncertainty in the value of a parameter
by placing a hyperprior on the variance of the Gaussian prior
on that parameter \citep{Berger:1985,Bishop:2006,Gelman:2013}:
\(
\sigma_{0}^{2} \given \alpha, \beta \dist \distInvGam(\alpha, \beta) 
\quad\text{and}\quad
\theta_{i} \given \mu_{0}, \sigma_{0}^{2} \dist \distNorm(\mu_{0}, \sigma_{0}^{2}),
\)
where $\distInvGam(\alpha, \beta)$ is the inverse gamma distribution with
shape $\alpha$ and scale $\beta$. Let $\nu \defined 2\alpha$ and $\sigma^{2} \defined \beta/\alpha$.
Then the marginal distribution of $\btheta$ follows the multivariate $t$-distribution
with location $\mu_{0}\bone$, scale matrix $\sigma^{2}I$, and $\nu$ degrees of freedom:
\(
\btheta \given \mu_{0}, \sigma^{2}, \nu \dist \distT_{\nu}(\mu_{0}\bone, \sigma^{2}I),
\)
where $\bone$ is the all-ones vector.
The multivariate $t$-distribution density is
\(
\lefteqn{p_{\distT}(\btheta \given \bmu, \Sigma, \nu)} \\
&= \frac{\Gamma(\frac{\nu + n}{2})\left(1 + \frac{1}{\nu}(\btheta - \bmu)^{\top}\Sigma^{-1}(\btheta - \bmu)\right)^{-\frac{\nu + n}{2}}}{\Gamma(\frac{\nu}{2})\pi^{n/2}\nu^{n/2}|\Sigma|^{1/2}}.
\)
When $\nu$ is finite, the multivariate $t$-distribution is heavy-tailed: 
the probability of $\|\btheta\|$ decreases at a polynomial rate as
$\|\btheta\| \to \infty$, compared to the exponential rate for a
multivariate Gaussian. 
For example $\nu = \sigma^{2} = 1$ and $\Sigma = \sigma^{2}$ gives
the multivariate Cauchy distribution. 
A multivariate Gaussian with covariance matrix $\Sigma$ is recovered 
by taking $\nu \to \infty$. 
Placing a multivariate $t$-distribution prior on $\btheta$
yields the following regret bound:
\bnthm[Multivariate $t$-distribution regret] \label{thm:mvt-regret}
If $\btheta \dist \distT_{\nu}(\bzero, \sigma^{2}I)$, then $\mcR(Z, \btheta^{*})$
is bounded by
\[
\begin{split}
R_{Bayes}^{mvt}(Z, \btheta^{*})
\defined 
  \frac{\nu + n}{2} \ln\left(1 + \frac{\|\btheta^{*}\|^{2}}{\nu\sigma^{2}}\right) \\
\phantom{\defined} + \frac{n}{2} \ln \left(\frac{(\nu + 1)(\nu + n)}{\nu^{2}} + \frac{Tc(\nu + 1)\sigma^{2}}{\nu n} \right).
\end{split}
\]
\enthm
\cref{thm:gaussian-regret} can be obtained as a special case of \cref{thm:mvt-regret}
by taking $\nu \to \infty$. 

Assume $\nu \ge 1$. 
If $\frac{\|\btheta^{*}\|^{2}}{\nu \sigma^{2}}$ is small, 
then 
\(
F(\|\btheta\|) 
\defined \frac{\nu + n}{2} \ln\left(1 + \frac{\|\btheta^{*}\|^{2}}{\nu\sigma^{2}}\right)
\approx \frac{n + \nu}{\nu}\frac{\|\btheta^{*}\|^{2}}{\sigma^{2}},
\)
so for ``small'' values of $\|\btheta^{*}\|^{2}$ (relative to $\nu \sigma^{2}$) the 
regret bound behaves similarly to having a Gaussian prior on $\btheta$.
However, if $\frac{\|\btheta^{*}\|^{2}}{\nu \sigma^{2}} \gg 1$, 
then the regret bound grows only logarithmically with $\|\btheta\|$, as we
would expect given that the multivariate $t$-distribution has heavy tails. 
Roughly speaking, $F(x)$
can be thought of as switching from quadratic to logarithmic behavior
when $x^{2} = \nu\sigma^{2}$, since this is the value 
at which $F$ switches from being convex to concave.

In general, the regret bound is large when the choice of 
$\btheta^{*}$ with small loss has large magnitude.
If a Gaussian prior is used, the possibility of 
$\|\btheta^{*}\|$ being large can be 
ameliorated by choosing $\sigma^{2}$ large,
since there is only a logarithmic regret penalty in $\sigma$. 
However, without a priori knowledge of how large the optimal
$\btheta^{*}$ might be, choosing a multivariate $t$-distribution
prior with a small value for $\nu$ and a moderate value for $\sigma^{2}$
allows for guaranteed logarithmic regret in the magnitude of
$\btheta^{*}$ no matter how large $\|\btheta^{*}\|$ is.
Hence, the use of the hierarchical (multivariate $t$-distribution) prior
does in fact yield greater robustness than the non-hierarchical 
(Gaussian) prior.\footnote{A more rigorous version of this statement
can be obtained for the Gaussian regression likelihood
by using the fact that there is a matching lower bound on the 
regret for the Gaussian prior/Gaussian regression model~\citep{Kakade:2004}.}

We can, in fact, develop more specific guidance on the choice of 
hyperparameters for the $t$-distribution. 
Our goal is to choose $\nu$ such that we obtain a $t$-distribution 
regret bound that is essentially as good as the Gaussian prior regret 
bound 
$R_{Bayes}^{G}
= \frac{\|\btheta^{*}\|^{2}}{2\sigma^{2}}
   + \frac{n}{2} \ln\left(1 + \frac{Tc\sigma^{2}}{n}\right)
$
for small $\|\btheta^{*}\|$ and better when 
$\|\btheta^{*}\|$ is large. 
If we choose $\nu$ equal to a constant, then for 
$n$ much larger than $\nu$, we have 
$R_{Bayes}^{mvt} \approx \frac{n}{2} \ln\left(1 + \frac{\|\btheta^{*}\|^{2}}{\nu\sigma^{2}}\right)
  + \frac{n}{2} \ln \left(\frac{n}{\nu} + \frac{Tc\sigma^{2}}{n} \right)$.
In the case of $\|\btheta^{*}\|$ small, 
we therefore have that the first term of $R_{Bayes}^{mvt}$ is approximately 
$\frac{n}{\nu}\frac{\|\btheta^{*}\|^{2}}{2\sigma^{2}}$,
and thus larger than the first term of $R_{Bayes}^{G}$ by a factor of $n/\nu$.
Furthermore, for small $T \ll \frac{n}{c \sigma^{2}}$ and any $\btheta^{*}$, the second term 
of $R_{Bayes}^{mvt}$ is approximately $\frac{n}{2}\ln(n/\nu)$ whereas
the second term of $R_{Bayes}^{G}$ is approximately $T c \sigma^{2} \ll n$. 
Thus, $R_{Bayes}^{mvt}$ with constant $\nu$ is not competitive with $R_{Bayes}^{G}$
in the large $n$ and small $T$ regimes. 
Instead consider the choice $\nu = Cn$ for constant $C > 0$, so
\(
\lefteqn{R_{Bayes}^{mvt}} \\
&\approx
  \frac{(C+1)n}{2} \ln\left(1 + \frac{\|\btheta^{*}\|^{2}}{Cn\sigma^{2}}\right)
  + \frac{n}{2} \ln \left(\frac{C+1}{C} + \frac{Tc\sigma^{2}}{n} \right) \\
&\le \frac{C+1}{C}\frac{\|\btheta^{*}\|^{2}}{2\sigma^{2}} 
  + \frac{n}{2} \ln \left(\frac{C+1}{C} + \frac{Tc\sigma^{2}}{n} \right).
\)
In this case, by choosing a moderate value of $C$, we see that a multivariate
$t$-distribution prior with $\nu = Cn$ has a competitive regret bound with a 
Gaussian prior in the small $\|\btheta^{*}\|$ regime, and 
exponentially smaller regret bound as $\|\btheta^{*}\|$ becomes large. 
Furthermore, the $t$-distribution prior remains competitive with the
Gaussian prior when $T$ is small. 

\subsection{Hierarchical Priors for Sharing Statistical Strength}
\label{sec:share-strength}

\subsubsection{Background}

We next consider hierarchical priors that allow for the 
sharing of statistical strength, providing answers to
\ref{q:sharing} and \ref{q:flat}: we specify some conditions
under which sharing of statistical strength can be achieved
and others in which a non-hierarchical prior is preferable.\footnote{For simplicity 
results are for Gaussian priors, though the 
extension to multivariate $t$-distribution priors is straightforward.}
In the machine learning literature, the goal of 
``sharing statistical strength'' has been formalized 
via  multitask learning (MTL) and 
``learning-to-learn'' (LTL) frameworks. 
A number of theoretical investigations of MTL and 
LTL haven been carried out, beginning with a series of papers
by Baxter~\citep[cf.][]{Baxter:1997,Baxter:2000}. 
Generically, such MTL and LTL frameworks involve two or more 
learning problems that are related to each other 
in some manner. 
The learning properties are investigated as the
number of tasks and/or the number of examples from
each task is increased. 
\citet{Baxter:2000} and \citet{BenDavid:2003} 
give sample complexity bounds based on classical
ideas from statistical and PAC learning theory. 
\citet{Baxter:1997} examines the asymptotic 
learning properties of hierarchical Bayesian
models. 
\citet{Pentina:2014} take a PAC-Bayesian approach while
\citet{HassanMahmud:2007,HassanMahmud:2009}, and 
\citet{Juba:2006} develop notions of task-relatedness 
from an (algorithmic) information-theoretic perspective.

Typically, tasks are equated with 
probability distributions over
examples (e.g., $(\bx, y)$ pairs). 
It is assumed that the tasks are drawn \iid\ from 
an unknown task distribution.
The goal is to learn the individual tasks
and learn about the task distribution.
Alternatively, a notion of
similarity can be used to relate the tasks:
the more similar the tasks, the greater the
advantage of learning them using multitask 
algorithms. 
In the online learning framework no 
assumptions are made about the distribution
of examples, so we consider two 
MTL scenarios in line with the latter setting. 
In the first scenario, one example from one
task is received at each time step. 
In the second, which is described in the 
\cref{sec:multi-obs}, at each time step
an example for each task is received 
simultaneously.
 
\subsubsection{Sequential Observations from Multiple Sources}

The sequential observation setting is relevant to the image
classification example given in the introduction, in which
there are many observations from 
some data sources and only a small number of observations 
from numerous other data sources.  
To model this situation, at  
time step $t$, an input $\bx_{t}$ from source $z_{t}$ is observed, 
where $z_{t} \in \theset{1,\dots,K}$. 
The learner predicts $y_{t}$ according to the posterior 
of $\btheta^{(z_{t})}$ given $Z_{t-1}$. 
An equivalent formulation is that the Bayesian learner observes
$\bx_{t} = (\bzero, \dots, \bx_{t}^{(k)}, \bzero, \dots)$ 
(if $z_{t} = k$) at each time, then receives $y_{t}$. 
Instead of using independent Gaussian priors 
on $\btheta^{(1)}, \dots,\btheta^{(K)}$,  
place a prior over the means of the $K$ priors. 
For each dimension $j = 1,\dots,n$, let
$
\mu_{j} \given \sigma_{0}^{2} \dist \distNorm(0, \sigma_{0}^{2})
$
and
$
\theta^{(k)}_{j} \given \mu_{j}, \sigma^{2} \dist \distNorm(\mu_{j}, \sigma^{2}), \quad k = 1,\dots,K,
$
and write
$\btheta_{j}^{(1:K)} \defined (\theta^{(1)}_{j}, \dots, \theta^{(K)}_{j})$.
Integrating out $\mu_{j}$ yields
\[
\btheta_{j}^{(1:K)} \given \sigma_{0}^{2}, \sigma^{2} \dist \distNorm(\bzero, \Sigma), 
\label{eq:gaussian-hierarchical-prior}
\]
where, with $1_{K}$ denoting the $K \times K$ all-ones matrix,
\[
\Sigma &\defined s^{2}\rho 1_{K} + s^{2}(1-\rho)I \\
s^{2} &\defined \sigma_{0}^{2} + \sigma^{2}, &
\rho &\defined \sigma_{0}^{2}/(\sigma_{0}^{2} + \sigma^{2}),
\]
This prior corresponds to the one-level prior in \citet{Salakhutdinov:2011}.
Similar results, which will be discussed qualitatively below,
can be obtained for the two-level prior at the cost
of a significantly more complicated bound. 
Define $T^{(k)} \defined \sum_{t=1}^{T} \delta_{k}(z_{t})$,
$Z^{(k)} \defined \{(x, y) \in Z | z_{t} = k \}$, and
$\gamma^{2} \defined K\sigma_{0}^{2} + \sigma^{2}$

\bnthm[Hierarchical Gaussian regret, sequential observations] \label{thm:hg-regret-sequential}
If $\btheta_{j}^{(1:K)} \dist \distNorm(\bzero, \Sigma)$,
$j = 1,\dots,n$, then $\mcR(Z, \btheta^{*})$ is bounded by
\[
&R_{Bayes}^{HG-seq}(Z, \btheta^{*}) 
\defined \frac{1}{2\gamma^{2}}\textstyle\sum_{k=1}^{K}\|\btheta^{*(k)}\|^{2}\nonumber \\ 
& + \frac{\sigma_{0}^{2}}{\sigma^{2}\gamma^{2}}\textstyle\sum_{k < \ell}\|\btheta^{*(k)} - \btheta^{*(\ell)}\|^{2} 
+ \frac{n}{2} \ln\left(1 + \frac{K\sigma_{0}^{2}}{\sigma^{2}}\right) \nonumber \\
&+ \frac{n}{2}\textstyle\sum_{k=1}^{K} \ln\left(1 - \frac{\sigma_{0}^{2}}{\gamma^{2}} + \frac{T^{(k)}c\sigma^{2}}{n}\right).
\label{eq:l-bayes-hg-seq}
\]
\enthm

It is instructive to compare 
the upper bound given in \eqref{eq:l-bayes-hg-seq}
to $\sum_{k} R_{Bayes}^{G}(Z_{(k)}, \btheta^{*(k)})$ with prior variance 
$s^{2} = \sigma_{0}^{2} + \sigma^{2}$. 
Setting $\sigma_{0} = \sigma$ 
yields a condition
for the hierarchical model to have 
smaller regret bound than the non-hierarchical 
model:
\[
\begin{split}
&4\|\btheta^{*(1)} - \btheta^{*(2)}\|^{2} + 3s^{2}n \textstyle\sum_{k=1}^{2}\ln\left(\frac{\frac{4}{3}n + T^{(i)}cs^{2}}{n + T^{(k)}cs^{2}}\right) \\
&\le \|\btheta^{*(1)}\|^{2} + \|\btheta^{*(2)}\|^{2} + 0.863 s^{2}n. \label{eq:h-nh-diff-seq}
\end{split}
\]

Of particular interest is the ``one-shot learning'' scenario, in which
only one observation (or a small number of observations) from a  
source are made while many observations are made from some other sources.
This setting is exactly that of the image classification problem of \citet{Salakhutdinov:2011}.
For concreteness, consider a ``large data'' task with $T^{(1)} \gg \frac{n}{cs^{2}}$
and a ``small data'' task $T^{(2)} = 2$, so that 
$\ln\left(\frac{\frac{4}{3}n + T^{(1)}cs^{2}}{n + T^{(1)}cs^{2}}\right) \approx 0$
and \eqref{eq:h-nh-diff-seq} becomes (approximately)
\(
&4\|\btheta^{*(1)} - \btheta^{*(2)}\|^{2} + 3s^{2}n \ln\left(\frac{4n + 6cs^{2}}{3n + 6cs^{2}}\right) \\
&\le \|\btheta^{*(1)}\|^{2} + \|\btheta^{*(2)}\|^{2} + 0.863 s^{2}n.
\)
But even for $n = 1$, 
$3 \ln\left(\frac{4n + 6cs^{2}}{3n + 6cs^{2}}\right) < 0.863$,
so the hierarchical model has smaller regret bound as long as 
$
4\|\btheta^{*(1)} - \btheta^{*(2)}\|^{2} 
\le \|\btheta^{*(1)}\|^{2} + \|\btheta^{*(2)}\|^{2} + Cs^{2}n
$
for some constant $0 < C < 0.863$. 

Hierarchical models for one-shot learning are designed with the goal
of providing good predictive power on the new problem (the second data source)
even with a small number of examples from that problem. 
To see if this is in fact the case for the hierarchical prior considered here,
we can investigate how much greater the regret bound is for 
$T^{(2)} > 0$ than for $T^{(2)} = 0$ with $T^{(1)} \gg \frac{n}{cs^{2}}$ fixed. 
With $\sigma_{0}^{2} = \sigma^{2}$, \eqref{eq:l-bayes-hg-seq} is greater 
in the former ($K=2$) than the latter ($K=1$) scenario by at most

\(
&- \frac{\|\btheta^{*(1)}\|^{2} }{6s^{2}}
+ \frac{\|\btheta^{*(2)}\|^{2}}{3s^{2}} 
+ \frac{2\|\btheta^{*(1)} - \btheta^{*(2)}\|^{2}}{3s^{2}}
+ \frac{3T^{(2)}cs^{2}}{8}.
\)
So if $\|\btheta^{*(2)}\|$ is small and $\btheta^{*(2)}$ and 
$\btheta^{*(1)}$ are close in $\ell^{2}$ distance,
the regret bound for the second source is small. 

The regret bound for the two-level prior in \citet{Salakhutdinov:2011}
is quite similar to that for the one-level prior.
Let $s_{k} \in \theset{1,\dots,S}$ denote the superclass of class $k$.
In the case of image classification, object classes that 
have similar visual appearance would have a common superclass.
The two-level prior consistes of an overall parameter prior
$\bbeta \dist \distNorm(0,\sigma_{0}^{2}I)$, superclass parameter priors
$\bmu^{(s)} \dist \distNorm(\bbeta,\sigma_{1}^{2}I)$, and class parameter priors
$\btheta^{(k)} \dist \distNorm(\bmu^{(s_{k})},\sigma_{2}^{2}I)$.
The regret bound for the two-level prior is
\[
\begin{split}
&c_{0}\textstyle\sum_{k=1}^{K}\|\btheta^{*(k)}\|^{2} 
+ \textstyle\sum_{k < \ell}c_{k\ell}\|\btheta^{*(k)} - \btheta^{*(\ell)}\|^{2} \\
&+ \frac{n}{2}\textstyle\sum_{k=1}^{K} O(\ln(c_{1} + c_{2}T^{(k)})) + O(1),
\end{split} \label{eq:l-bayes-hg-seq-2}
\]
where $c_{0}, c_{1}, c_{2} > 0$ are constants, $c_{k\ell} = \tc_{s_{k}}$ if
$s_{k} = s_{\ell}$ and $c_{k\ell} = \tc_{s_{k}s_{\ell}}$ if $s_{k} \ne s_{\ell}$. 
Furthermore, $\tc_{s} > \tc_{s's''}$ for all $s,s',s'' \in \theset{1,\dots,S}$. 
Hence, the regret bound's being small depends more on the parameter 
vectors in the same superclass being close to each other
than on parameter vectors from different superclasses being
close to each other. 
See \cref{sec:two-level-prior} for details. 
The two-level regret bound well-explains the results of \citet{Salakhutdinov:2011}.
The poor performance on image classes with very different 
visual appearance from the other classes is unsurprising since the
parameter vectors that predict these classes well are going to have
large $\ell^{2}$ distance from the parameter vectors of other 
object classes.

\subsection{Hierarchical Priors for Feature Selection}
\label{sec:sparsity}

A shortcoming of the priors investigated so far is the poor dependence on
the feature space dimension $n$. 
For example, the Gaussian prior 
regret bound is (approximately)
\(
L_{Bayes}(Z) 
\le \inf_{\btheta^{*}} L_{\btheta^{*}}(Z) + \frac{1}{2\sigma^{2}}\|\btheta^{*}\|^{2} 
   + \frac{Tc\sigma^{2}}{2},
\)
when $Tc\sigma^{2} \ll n$, so the regret may grow
linearly in this regime.\footnote{Dimension-independent 
regret bounds for the priors already considered can be obtained,
but at the price of a constant greater than one front of 
the $L_{\btheta^{*}}(Z)$ term. See, e.g., \citet{Banerjee:2007}.}
In the infinite-dimensional case, Gaussian processes can be used while
still obtaining meaningful regret bounds~\citep{Kakade:2005,Seeger:2008}.
However, methods that are applicable to high-dimensional problems
for which $n \gg T$ but $n$ is still finite are of great general interest. 
For example, in the image classification example from the introduction, the
feature vector has $n \approx 5000$, whereas 
most object classes have fewer than 200 training examples. 
In high-dimensional problems it is desirable to use feature selection or
sparse methods to reduce the effective dimension of the 
problem, with the aim of achieving better generalization performance 
and increasing interpretability of the model. 
A popular non-Bayesian approach for inducing sparsity is $\ell_{1}$ 
regularization, such as the lasso
for linear regression~\citep{Tibshirani:1996}.
A Bayesian approach is the Bayesian lasso: the $\ell_{1}$ 
regularizer of the lasso is converted into a prior, which amounts to placing 
a Laplace prior on $\btheta$~\citep{Park:2008}.
However, the Bayesian lasso still seems to lead to a linear dependence 
on the dimension 
because the model
puts zero prior mass on a component being exactly zero.
A regret bound for the Bayesian lasso can be found in the 
\cref{sec:bayesian-lasso} (we suspect that our bound is essentially tight, 
though we have been unable to obtain a matching lower bound). 

Another common Bayesian approach to inducing sparsity 
is to use a hierarchical ``spike and slab'' prior,
which places positive probability on a component
being exactly zero~\citep{Ishwaren:2005,Narisetty:2014}.
One version of the spike and slab prior is
\(
z_{i} \given p \dist \distBern(p) 
\quad \text{and} \quad
\theta_{i} \given z_{i} \dist z_{i}\delta_{0} + (1-z_{i})\distNorm(0, \sigma^{2}).
\)
So with probability $p$ component $i$ is zero and 
with probability $1-p$ it is Gaussian-distributed.
Integrating out $z_{i}$ yields prior density
$
p_{0}(\theta_{i}) = p\delta_{0}(\theta_{i}) + (1-p)\distNorm(\theta_{i} \given 0, \sigma^{2}).
$
Let $\|\bv\|_{0}$ denote the $\ell_{0}$ norm of the vector $\bv$. 
\bnthm \label{thm:ss-regret}
For the spike-and-slab prior, if $m = \|\btheta^{*}\|_{0}$,
then $\mcR(Z, \btheta^{*})$ is bounded by
\[
R_{Bayes}^{SS}(Z, \btheta^{*})
&\defined 
   \frac{1}{2\sigma^{2}}\|\btheta^{*}\|^{2} 
   + m \ln \frac{1}{1-p} \label{eq:ss-regret-bound} \\
   &+ (n - m) \ln \frac{1}{p}
   + \frac{m}{2} \ln\left(1 + \frac{Tc\sigma^{2}}{m}\right).  \nonumber
\]
In particular, if $p \defined q^{1/n}$ for some constant $0 < q < 1$, then 
$R_{Bayes}^{SS}(Z, \btheta^{*})$  is at most
\[
     \frac{\|\btheta^{*}\|^{2} }{2\sigma^{2}}
   + m \ln \frac{n}{1-q} + \ln \frac{1}{q}
   + \frac{m}{2} \ln\Big(1 + \frac{Tc\sigma^{2}}{m}\Big). \label{eq:ss-regret-bound-2}
\]
\enthm
The theorem shows the importance of properly scaling $p$ with the dimension
of the problem. 
If $p$ is kept fixed, then the regret has linear dependence on $n$. 
However, by scaling $p$ to be $q^{1/n}$, we increase the probability of a 
component being zero as the dimension increases
and thus are able to ensure that the regret is only logarithmic in $n$
while simultaneously maintaining the appropriate linear dependence on $m$. 
The constant $q$ turns out to be the prior probability that all of the components are set to zero. 
More generally, 
${n \choose k} q^{(n-k)/n}(1-q^{1/n})^{k} \overset{n \to \infty}{\to} \frac{q \ln^{k} q^{-1}}{k!}$,
the limiting prior probability of choosing exactly $k$ components to be non-zero.
Hence, as $n \to \infty$, the prior over the number of non-zero components converges 
to a Poisson distribution with rate parameter $\ln q^{-1}$.  
So when $n$ is large the expected number of non-zero components is $\approx \ln q^{-1}$. 
The choice of $p$ close to 1 for large $n$ is in notable 
contrast to the common practice of setting $p = \half$ or 
some other constant independent of $n$ 
\citep{Schneider:2004,Ishwaren:2005}. 
Our results strongly recommend against this practice. 
See \citet{Narisetty:2014} for a discussion of purely 
statistical reasons to scale $p$ with the dimension.

\section{Conclusion}

In this paper we set out to understand and quantify the learning-theoretic
benefits of Bayesian hierarchical modeling. 
In \cref{sec:applications}, we used first our main result, \cref{thm:meta-theorem}, to 
analyze three specific hierarchical priors that, 
particularly when combined with a logistic or Gaussian regression likelihood, 
are widely used in practice.
Indeed, these prior-likelihood combinations have often been used with substantial
success even in situations when they are known to be rather poor models for the data generating mechanism. 
Our analysis offers an explanation for this success. 
The priors we analyzed are representative of the variety of ways in which hierarchical models are employed: 
representing uncertainty in hyperparameters, tying together related groups of observations, 
and creating more complicated distributions from simpler ones.
Thus, our results answer Questions \ref{q:sharing}-\ref{q:hypers} in
some important cases and exemplify a learning-theoretic analysis technique that can be
applied to other hierarchical models. 
In addition, using our second main result, \cref{thm:pac-bayes-regret}, all of the insights
gained in \cref{sec:applications} for the log-loss regret setting 
apply equally well to the batch setting of statistical risk with bounded loss,
further extending the applicability of our conclusions. 

\newpage

\section*{Acknowledgments}
Thanks to Peter Gr\"unwald, Sham Kakade, Peter Krafft, and Daniel Roy
for helpful discussions and comments.
Thanks also to the anonymous reviewers whose constructive comments
improved the presentation of the results. 
JHH was supported by the U.S. Government under FA9550-11-C-0028 and awarded by the DoD, Air Force
Office of Scientific Research, National Defense Science and Engineering Graduate
(NDSEG) Fellowship, 32 CFR 168a.

\bibliography{bayes-regret}

\begin{thebibliography}{35}
\providecommand{\natexlab}[1]{#1}
\providecommand{\url}[1]{\texttt{#1}}
\expandafter\ifx\csname urlstyle\endcsname\relax
  \providecommand{\doi}[1]{doi: #1}\else
  \providecommand{\doi}{doi: \begingroup \urlstyle{rm}\Url}\fi

\bibitem[Alkema et~al.(2011)Alkema, Raftery, Gerland, Clark, Pelletier,
  Buettner, and Heilig]{Alkema:2011}
Alkema, L., Raftery, A.~E., Gerland, P., Clark, S.~J., Pelletier, F., Buettner,
  T., and Heilig, G.~K.
\newblock {Probabilistic Projections of the Total Fertility Rate for All
  Countries}.
\newblock \emph{Demography}, 48\penalty0 (3):\penalty0 815--839, July 2011.

\bibitem[Audibert \& Bousquet(2007)Audibert and Bousquet]{Audibert:2007}
Audibert, J.-Y. and Bousquet, O.
\newblock {Combining PAC-Bayesian and generic chaining bounds}.
\newblock \emph{The Journal of Machine Learning Research}, 8:\penalty0
  863--889, 2007.

\bibitem[Banerjee(2006)]{Banerjee:2006}
Banerjee, A.
\newblock {On Bayesian Bounds}.
\newblock In \emph{International Conference on Machine Learning}, pp.\  81--88.
  ACM, 2006.

\bibitem[Banerjee(2007)]{Banerjee:2007}
Banerjee, A.
\newblock {An Analysis of Logistic Models: Exponential Family Connections and
  Online Performance}.
\newblock In \emph{International Conference on Data Mining}, pp.\  204--215,
  2007.

\bibitem[Baxter(1997)]{Baxter:1997}
Baxter, J.
\newblock {A Bayesian/information theoretic model of learning to learn via
  multiple task sampling}.
\newblock \emph{Machine learning}, 28\penalty0 (1):\penalty0 7--39, 1997.

\bibitem[Baxter(2000)]{Baxter:2000}
Baxter, J.
\newblock {A Model of Inductive Bias Learning}.
\newblock \emph{Journal of Artificial Intelligence Research}, 12:\penalty0
  149--198, 2000.

\bibitem[Ben-David \& Schuller(2003)Ben-David and Schuller]{BenDavid:2003}
Ben-David, S. and Schuller, R.
\newblock {Exploiting task relatedness for multiple task learning}.
\newblock In \emph{Conference on Learning Theory}, pp.\  567--580. Springer,
  2003.

\bibitem[Berger(1985)]{Berger:1985}
Berger, J.~O.
\newblock \emph{{Statistical Decision Theory and Bayesian Analysis}}.
\newblock Springer-Verlag, New York, second edition, 1985.

\bibitem[Bernardo \& Smith(2000)Bernardo and Smith]{BS:2000}
Bernardo, J.~M. and Smith, A. F.~M.
\newblock \emph{{Bayesian Theory}}.
\newblock Wiley, New York, 2000.

\bibitem[Bishop(2006)]{Bishop:2006}
Bishop, C.~M.
\newblock \emph{{Pattern Recognition and Machine Learning}}.
\newblock Springer, 2006.

\bibitem[Catoni(2007)]{Catoni:2007}
Catoni, O.
\newblock \emph{{PAC-Bayesian Supervised Classification: The Thermodynamics of
  Statistical Learning}}, volume~56 of \emph{Lecture Notes - Monograph Series}.
\newblock Institute of Mathematical Statistics, 2007.

\bibitem[Cesa-Bianchi \& Lugosi(2006)Cesa-Bianchi and Lugosi]{CesaBianchi:2006}
Cesa-Bianchi, N. and Lugosi, G.
\newblock \emph{{Prediction, Learning, and Games}}.
\newblock Cambridge University Press, New York, 2006.

\bibitem[Dawid \& Vovk(1999)Dawid and Vovk]{Dawid:1999}
Dawid, A.~P. and Vovk, V.~G.
\newblock {Prequential probability: Principles and properties}.
\newblock \emph{Bernoulli}, 5\penalty0 (1):\penalty0 125--162, 1999.

\bibitem[Gelman \& Hill(2006)Gelman and Hill]{Gelman:2006}
Gelman, A. and Hill, J.
\newblock \emph{{Data Analysis Using Regression and Multilevel/Hierarchical
  Models}}.
\newblock Cambridge University Press, 2006.

\bibitem[Gelman et~al.(2013)Gelman, Carlin, Stern, Dunson, Vehtari, and
  Rubin]{Gelman:2013}
Gelman, A., Carlin, J., Stern, H., Dunson, D., Vehtari, A., and Rubin, D.~B.
\newblock \emph{{Bayesian Data Analysis}}.
\newblock Chapman and Hall/CRC, third edition, 2013.

\bibitem[Ghitza \& Gelman(2013)Ghitza and Gelman]{Ghitza:2013}
Ghitza, Y. and Gelman, A.
\newblock {Deep Interactions with MRP: Election Turnout and Voting Patterns
  Among Small Electoral Subgroups}.
\newblock \emph{American Journal of Political Science}, 57\penalty0
  (3):\penalty0 762--776, February 2013.

\bibitem[Gr{\"u}nwald(2007)]{Grunwald:2007}
Gr{\"u}nwald, P.~D.
\newblock \emph{{The Minimum Description Length Principle}}.
\newblock MIT Press, Cambridge, MA, 2007.

\bibitem[Hassan~Mahmud(2009)]{HassanMahmud:2009}
Hassan~Mahmud, M.~M.
\newblock {On universal transfer learning}.
\newblock \emph{Theoretical Computer Science}, 410\penalty0 (19):\penalty0
  1826--1846, 2009.

\bibitem[Hassan~Mahmud \& Ray(2007)Hassan~Mahmud and Ray]{HassanMahmud:2007}
Hassan~Mahmud, M.~M. and Ray, S.~R.
\newblock {Transfer Learning using Kolmogorov Complexity: Basic Theory and
  Empirical Evaluations}.
\newblock In \emph{Advances in Neural Information Processing Systems}, pp.\
  361--368, 2007.

\bibitem[Ishwaran \& Rao(2005)Ishwaran and Rao]{Ishwaren:2005}
Ishwaran, H. and Rao, J.~S.
\newblock {Spike and slab variable selection: Frequentist and Bayesian
  strategies}.
\newblock \emph{The Annals of Statistics}, 33\penalty0 (2):\penalty0 730--773,
  April 2005.

\bibitem[Juba(2006)]{Juba:2006}
Juba, B.
\newblock {Estimating Relatedness via Data Compression}.
\newblock In \emph{International Conference on Machine Learning}, pp.\
  441--448, 2006.

\bibitem[Kakade \& Ng(2004)Kakade and Ng]{Kakade:2004}
Kakade, S.~M. and Ng, A.~Y.
\newblock {Online Bounds for Bayesian Algorithms}.
\newblock In \emph{Advances in Neural Information Processing Systems}, 2004.

\bibitem[Kakade et~al.(2005)Kakade, Seeger, and Foster]{Kakade:2005}
Kakade, S.~M., Seeger, M., and Foster, D.~P.
\newblock {Worst-case bounds for Gaussian process models}.
\newblock In \emph{Advances in Neural Information Processing Systems}, pp.\
  193--200, 2005.

\bibitem[McAllester(2003)]{McAllester:2003a}
McAllester, D.~A.
\newblock {Simplified PAC-Bayesian Margin Bounds}.
\newblock In \emph{Conference on Learning Theory}, pp.\  203--215, 2003.

\bibitem[Mengersen \& Robert(2014)Mengersen and Robert]{Mengersen:2014}
Mengersen, K.~L. and Robert, C.~P.
\newblock {Big Bayes Stories---Foreword}.
\newblock \emph{Statistical Science}, 29\penalty0 (1):\penalty0 1--1, 2014.

\bibitem[Narisetty \& He(2014)Narisetty and He]{Narisetty:2014}
Narisetty, N.~N. and He, X.
\newblock {Bayesian variable selection with shrinking and diffusing priors}.
\newblock \emph{The Annals of Statistics}, 42\penalty0 (2):\penalty0 789--817,
  April 2014.

\bibitem[Park \& Casella(2008)Park and Casella]{Park:2008}
Park, T. and Casella, G.
\newblock {The Bayesian Lasso}.
\newblock \emph{Journal of the American Statistical Association}, 103\penalty0
  (482):\penalty0 681--686, June 2008.

\bibitem[Pentina \& Lampert(2014)Pentina and Lampert]{Pentina:2014}
Pentina, A. and Lampert, C.~H.
\newblock {A PAC-Bayesian bound for Lifelong Learning}.
\newblock In \emph{International Conference on Machine Learning}, 2014.

\bibitem[Raftery et~al.(2012)Raftery, Li, {\v S}ev{\v c}{\'\i}kov{\'a},
  Gerland, and Heilig]{Raftery:2012}
Raftery, A.~E., Li, N., {\v S}ev{\v c}{\'\i}kov{\'a}, H., Gerland, P., and
  Heilig, G.~K.
\newblock {Bayesian probabilistic population projections for all countries}.
\newblock \emph{Proceedings of the National Academy of Sciences}, 109\penalty0
  (35):\penalty0 13915--13921, 2012.

\bibitem[Raftery et~al.(2013)Raftery, Chunn, Gerland, and {\v S}ev{\v
  c}{\'\i}kov{\'a}]{Raftery:2013}
Raftery, A.~E., Chunn, J.~L., Gerland, P., and {\v S}ev{\v c}{\'\i}kov{\'a}, H.
\newblock {Bayesian Probabilistic Projections of Life Expectancy for All
  Countries}.
\newblock \emph{Demography}, 50\penalty0 (3):\penalty0 777--801, March 2013.

\bibitem[Salakhutdinov et~al.(2011)Salakhutdinov, Torralba, and
  Tenenbaum]{Salakhutdinov:2011}
Salakhutdinov, R., Torralba, A., and Tenenbaum, J.~B.
\newblock {Learning to share visual appearance for multiclass object
  detection}.
\newblock In \emph{Conference on Computer Vision and Pattern Recognition}, pp.\
   1481--1488. IEEE, 2011.

\bibitem[Schneider \& Corcoran(2004)Schneider and Corcoran]{Schneider:2004}
Schneider, U. and Corcoran, J.~N.
\newblock {Perfect sampling for Bayesian variable selection in a linear
  regression model}.
\newblock \emph{Journal of Statistical Planning and Inference}, 126\penalty0
  (1):\penalty0 153--171, November 2004.

\bibitem[Seeger et~al.(2008)Seeger, Kakade, and Foster]{Seeger:2008}
Seeger, M.~W., Kakade, S.~M., and Foster, D.~P.
\newblock {Information Consistency of Nonparametric Gaussian Process Methods}.
\newblock \emph{Information Theory, IEEE Transactions on}, 54\penalty0
  (5):\penalty0 2376--2382, 2008.

\bibitem[Tibshirani(1996)]{Tibshirani:1996}
Tibshirani, R.
\newblock {Regression shrinkage and selection via the lasso}.
\newblock \emph{Journal of the Royal Statistical Society: Series B (Statistical
  Methodology)}, pp.\  267--288, 1996.

\bibitem[Vovk(2001)]{Vovk:2001}
Vovk, V.
\newblock {Competitive On‐line Statistics}.
\newblock \emph{International Statistical Review}, 69\penalty0 (2):\penalty0
  213--248, 2001.

\end{thebibliography}
\bibliographystyle{icml2015}

\newpage
\onecolumn

\appendix 

\numberwithin{equation}{section}

\section{Regret Bounds for Non-GLM Likelihoods}

Recall Proposition \ref{prop:compression}, restated here for convenience:
\bprop 
The Bayesian cumulative loss is bounded as
\[
L_{Bayes}(Z_{T}) \le L_{Q}(Z_{T}) + \kl{Q}{P_{0}}. \label{eq:app-compression}
\]
\eprop

\bprf[Proof of \cref{thm:super-meta-theorem}]
Fix a choice of $\btheta^{*}$ and $\bphi$ and write $Q = Q_{\btheta^{*},\bphi}$. 
Take a second-order Taylor expansion of $f_{y}$ about $\bz^{*}$, yielding
\(
f_{y}(\bz) 
= f_{y}(\bz^{*}) + f_{y}'(\bz^{*})^{\top}(\bz - \bz^{*}) + \frac{1}{2}(\bz - \bz^{*})^{\top}f_{y}''(\bzeta(\bz))(\bz - \bz^{*}),
\)
for some function $\bzeta$. 
Let $\bz = (\bxi\bx, \bpsi)$ with $\btheta \dist Q$ and 
let $\bz^{*} = \EE[\bz] = (\bxi^{*}\bx, \bpsi^{*})$. 
Hence,
\(
\EE_{\bz}[f_{y}(\bz)]  
&= f_{y}(\bz^{*}) + f_{y}'(\bz^{*})^{\top} \bzero + \frac{1}{2}\EE_{\bz}\left[(\bz - \bz^{*})^{\top}f_{y}''(\bzeta(\bz))(\bz - \bz^{*})\right] \\
&\le f_{y}(\bz^{*}) + \frac{c}{2}\,\EE_{\bz}\left[(\bz - \bz^{*})^{\top}(\bz - \bz^{*})\right].
\)
Defining
\(
\bomega \defined (\underbrace{\bx,\dots,\bx}_\text{$n'$ times},\underbrace{1,\dots,1}_\text{$n''$ times}),
\)
we next observe that
\[
(\bz - \bz^{*})^{\top}(\bz - \bz^{*})
&= \bomega^{\top}(\btheta-\btheta^{*})(\btheta-\btheta^{*})^{\top}\bomega.
\]
Letting $\Sigma = \var[\btheta]$, we thus have
\(
\EE_{\bz}\left[(\bz - \bz^{*})^{\top}(\bz - \bz^{*})\right] 
&= \bomega^{\top}\EE_{\btheta}[(\btheta-\btheta^{*})(\btheta-\btheta^{*})^{\top}]\bomega \\
&\le \|\bomega\|_{2}^{2}\|\EE_{\btheta}[(\btheta-\btheta^{*})(\btheta-\btheta^{*})^{\top}]\| \\
&= (n'\|\bx\|^{2}_{2} + n'')\|\Sigma\| \\
&\le (n' + n'')\|\Sigma\|
\)
since it is assumed that $\|\bx\|_{2} \le 1$. 
Noting that $L_{Q}(Z_{T}) = \sum_{t} \EE_{Q}[f_{y_{t}}(\bxi\bx_{t},\bpsi)]$
and $L_{\btheta^{*}}(Z_{T}) = \sum_{t} f_{y_{t}}(\bxi^{*}\bx_{t},\bpsi^{*})$,
we have
\[
L_{Q}(Z_{T}) \le L_{\btheta^{*}}(Z_{T}) + \frac{Tc(n' + n'')\|\Sigma\|}{2}. \label{eq:lq-bound}
\]
Combining \eqref{eq:app-compression} and \eqref{eq:lq-bound} yields the theorem.
\eprf
\bprf[Proof of \cref{thm:meta-theorem}]
Follows as a special case of \cref{thm:super-meta-theorem} by choosing $n'=1$ and $n''=0$. 
\eprf

\subsection{Application to Multi-class Logistic Regression}

For multi-class logistic regression (MLR) $y \in \{1,\dots,K\}$ is one of $K$ classes,
the parameters are $\btheta = \{ \btheta^{(k)} \}_{k=1}^{K}$, and the likelihood is
\[
p(y \given \btheta, \bx) = \frac{\exp(\btheta^{(y)}\cdot \bx)}{\sum_{k=1}^{K}\exp(\btheta^{(k)}\cdot \bx)}.
\]
In order to apply \cref{thm:super-meta-theorem}, we require the following result: 
\bnprop
Assumption \eqref{eq:a1'} holds for the MLR likelihood with $c=\half$. 
\enprop
\bprf
First note that 
\[
f_{y}(\bz) = -z_{y} + \ln \textstyle\sum_{k=1}^{K}e^{z_{i}},
\]
where $z_{i} = \btheta^{(k)} \cdot \bx$,
and hence the Hessian of $f_{y}(\bz)$ is independent of $y$:
\[
f_{y}''(\bz) = \frac{1}{(\sum_{k=1}^{K}e^{z_{i}})^{2}}
\bmat
\sum_{i\ne1}e^{z_{1}+z_{i}} & -e^{z_{1}+z_{2}} & \hdots & -e^{z_{1}+z_{K}} \\
-e^{z_{2}+z_{1}} & \sum_{i\ne2}e^{z_{2}+z_{i}} & \hdots & -e^{z_{2}+z_{K}}\\
\vdots & & \ddots 
\emat
\]
Applying Gershgorin's circle theorem, we find that 
\[
\|f_{y}''(\bz)\| \le \frac{2e^{z_{1}}\sum_{i\ne1}e^{z_{i}}}{(\sum_{k=1}^{K}e^{z_{k}})^{2}},
\]
where with loss of generality we have applied the theorem to the first 
row of the Hessian. 
Defining $a \defined e^{z_{1}} \ge 0$ and $b \defined \sum_{i\ne1}e^{z_{i}} \ge 0$, we 
have $\|f_{y}''(\bz)\| \le \frac{2ab}{(a+b)^{2}}$. 
Maximization over the positive orthant occurs at $a = b > 0$, so 
$\|f_{y}''(\bz)\| \le \half$. 
\eprf

Reasoning similarly to \cref{thm:hg-regret-simultaneous}, one can 
easily prove:
\bnthm[Hierarchical Gaussian regret, multi-class regression] \label{thm:hg-mlr-regret}
If $\btheta_{j}^{(1:K)} \dist \distNorm(\bzero, \Sigma)$, 
$j = 1,\dots,n$, then using the MLR likelihood guarantees that $\mcR(Z, \btheta^{*})$ is
bounded by
\[
\begin{split}
R_{Bayes}^{mlr-HG}(Z, \btheta^{*})
&\defined
   \frac{1}{2\gamma^{2}}\textstyle\sum_{k=1}^{K}\|\btheta^{*(k)}\|^{2} 
   + \frac{\sigma_{0}^{2}}{\sigma^{2}\gamma^{2}}\textstyle\sum_{k < \ell}\|\btheta^{*(k)} - \btheta^{*(\ell)}\|^{2} \\
&\phantom{\defined~} + \frac{n}{2} \ln\left(1 + \frac{K\sigma_{0}^{2}}{\sigma^{2}}\right) 
   + \frac{nK}{2} \ln\left(1 - \frac{\sigma_{0}^{2}}{\gamma^{2}} + \frac{T\sigma^{2}}{2n}\right),
\end{split} \label{eq:l-bayes-hg-mlr}
\]
where $\gamma^{2} \defined K\sigma_{0}^{2} + \sigma^{2}$.
\enthm

\cref{thm:mlr-gaussian-regret} follows as a special case of \cref{thm:hg-mlr-regret} 
by taking $\sigma_{0}^{2} = 0$.

\section{Proof of \cref{thm:pac-bayes-regret}}

Since 
$p_{T}(\btheta) = \frac{p(Y \given X, \btheta)p_{0}(\btheta)}{p(Y\given X)}$,
\[
\kl{P_{T}}{P_{0}}
&= \EE_{P_{T}}\left[\ln \frac{p_{T}(\btheta)}{p_{0}(\btheta)}\right] \nonumber \\
&= \EE_{P_{T}}\left[\ln \frac{p(Y\given X,\btheta)}{p(Y \given X)}\right] \nonumber \\
&= L_{Bayes}(Z_{T}) - L_{P_{T}}(Z_{T}). \label{eq:kl-PT-P0}
\]
Combining \eqref{eq:generic-regret-bound} and \eqref{eq:kl-PT-P0} 
with \cref{thm:pac-bayes} implies that with probability $1 - \delta$,
for all $\btheta$,
\(
|\mcL(P_{T}) - \hat\mcL(P_{T}, Z_{T})|
&\le \sqrt\kappa\sqrt{\frac{L_{\btheta}(Z_{T}) - L_{P_{T}}(Z_{T}) + B(\btheta) + C(T) + \ln \kappa'/\delta}{T}}. 
\)
Observing that $L_{\btheta^{*}}(Z_{T}) < L_{P_{T}}(Z_{T})$,
so $L_{\btheta^{*}}(Z_{T}) - L_{P_{T}}(Z_{T}) < 0$, completes the proof.

\section{KL Divergence Derivations}

\subsection{Multivariate Gaussians}
\label{app:gaussians-kl}

Let $D_{i} = \distNorm(\mu_{i}, \Sigma_{i}), i = 1,2$, where $\dim(\mu_{i}) = n$.
Then 
\(
\kl{D_{1}}{D_{2}} 
&= \frac{1}{2} \EE_{D_{1}}\left[\ln\frac{|\Sigma_{2}|}{|\Sigma_{1}|} - (x - \mu_{1})^{\top}\Sigma_{1}^{-1}(x - \mu_{1})
 + (x - \mu_{2})^{\top}\Sigma_{2}^{-1}(x - \mu_{2})\right] \\
&= \frac{1}{2}\left\{ \ln\frac{|\Sigma_{2}|}{|\Sigma_{1}|} + \EE_{D_{1}}\left[ - \tr(\Sigma_{1}^{-1}(x - \mu_{1})^{\top}(x - \mu_{1}))
 + \tr(\Sigma_{2}^{-1}(x - \mu_{2})^{\top}(x - \mu_{2}))\right]\right\} \\
&= \frac{1}{2}\left\{ \ln\frac{|\Sigma_{2}|}{|\Sigma_{1}|} - \tr(\Sigma_{1}^{-1}\Sigma_{1})
 + \EE_{D_{1}}\left[\tr(\Sigma_{2}^{-1}(x^{\top}x - 2x^{\top}\mu_{2} + \mu_{2}^{\top}\mu_{2}))\right]\right\} \\
&= \frac{1}{2}\left\{ \ln\frac{|\Sigma_{2}|}{|\Sigma_{1}|} - n
 + \EE_{D_{1}}\left[\tr(\Sigma_{2}^{-1}(x^{\top}x - 2x^{\top}\mu_{2} + \mu_{2}^{\top}\mu_{2}))\right]\right\} \\
&= \frac{1}{2}\left\{ \ln\frac{|\Sigma_{2}|}{|\Sigma_{1}|} - n
 + \tr(\Sigma_{2}^{-1}(\Sigma_{1} + \mu_{1}^{\top}\mu_{1} - 2\mu_{1}^{\top}\mu_{2} + \mu_{2}^{\top}\mu_{2}))\right\} \\
&= \frac{1}{2}\left\{ \ln\frac{|\Sigma_{2}|}{|\Sigma_{1}|}  - n + \tr(\Sigma_{2}^{-1}\Sigma_{1})
 + (\mu_{1} - \mu_{2})^{\top}\Sigma_{2}^{-1}(\mu_{1} - \mu_{2}) \right\}.
\)

\subsection{Gaussian and $t$-Distribution}
\label{app:gaussian-t-kl}

Let $D_{1} = \distNorm(\mu_{1}, \Sigma_{1})$ and
$D_{2} = \distT_{\nu}(\mu_{2}, \Sigma_{2})$, where $\dim(\mu_{i}) = k$.
Then
\(
\kl{D_{1}}{D_{2}}
&= \ln\left(\frac{\Gamma(\frac{\nu}{2})\nu^{k/2}}{\Gamma(\frac{\nu + k}{2})}\right) 
 + \frac{k}{2}\ln \pi + \frac{1}{2}\ln |\Sigma_{2}|  
 - \frac{k}{2} \ln 2\pi e - \frac{1}{2} \ln|\Sigma_{1}| \\
&\phantom{=~} 
 + \frac{\nu + k}{2} \EE_{D_{1}}\left[\ln\left(1 + \frac{1}{\nu}(x-\mu_{2})^{\top}\Sigma_{2}^{-1}(x-\mu_{2})\right)\right] \\
&= \ln\left(\frac{\Gamma(\frac{\nu}{2})\nu^{k/2}}{\Gamma(\frac{\nu + k}{2})}\right) 
 + \frac{1}{2}\ln\frac{|\Sigma_{2}|}{|\Sigma_{1}|}  - \frac{k}{2} \ln 2 e \\
&\phantom{=~} 
 + \frac{\nu + k}{2} \EE_{D_{1}}\left[\ln\left(1 + \frac{1}{\nu}(x-\mu_{2})^{\top}\Sigma_{2}^{-1}(x-\mu_{2})\right)\right].
\)
For the first term, if $k$ is even, then 
\(
\frac{\Gamma(\frac{\nu}{2})\nu^{k/2}}{\Gamma(\frac{\nu + k}{2})}
&= \frac{\nu^{k/2}}{(\frac{\nu + k}{2})^{\underline{k/2}}},
\)
where $y^{\underline{n}} = y(y-1)\dots(y-n+1)$ is the descending factorial. 
Now assume $k$ is odd. By Gautschi's inequality,
$\frac{\Gamma(a)}{\Gamma(a+1/2)} \le \left(\frac{2a + 1}{2a^{2}}\right)^{1/2}$.
Choosing $a = \nu/2$ yields
\(
\frac{\Gamma(\frac{\nu}{2})\nu^{k/2}}{\Gamma(\frac{\nu + k}{2})}
&= \frac{\Gamma(\frac{\nu}{2})\nu^{1/2}\nu^{(k-1)/2}}{\Gamma(\frac{\nu + 1}{2})(\frac{\nu + k}{2})^{\underline{(k-1)/2}}}
\le \frac{(\nu + 1)^{1/2}\nu^{(k-1)/2}}{(\frac{\nu}{2})^{1/2}(\frac{\nu + k}{2})^{\underline{(k-1)/2}}}.
\)
Now, bounding the expectation gives
\(
\lefteqn{\EE_{D_{1}}\left[\ln\left(1 + \frac{1}{\nu}(x-\mu_{2})^{\top}\Sigma_{2}^{-1}(x-\mu_{2})\right)\right]} \\
&\le \ln\left(1 + \frac{1}{\nu}\EE_{D_{1}}\left[(x-\mu_{2})^{\top}\Sigma_{2}^{-1}(x-\mu_{2})\right]\right) \\
&= \ln\left(1 + \frac{1}{\nu}\tr(\Sigma_{2}^{-1}\Sigma_{1})
 + \frac{1}{\nu}(\mu_{1} - \mu_{2})^{\top}\Sigma_{2}^{-1}(\mu_{1} - \mu_{2})\right) \\
&\le \ln\left(1 + \frac{1}{\nu}(\mu_{1} - \mu_{2})^{\top}\Sigma_{2}^{-1}(\mu_{1} - \mu_{2})\right) + \frac{\tr(\Sigma_{2}^{-1}\Sigma_{1})}{\nu + (\mu_{1} - \mu_{2})^{\top}\Sigma_{2}^{-1}(\mu_{1} - \mu_{2})} \\
&\le \ln\left(1 + \frac{1}{\nu}(\mu_{1} - \mu_{2})^{\top}\Sigma_{2}^{-1}(\mu_{1} - \mu_{2})\right) + \frac{1}{\nu}\tr(\Sigma_{2}^{-1}\Sigma_{1}),
\)
where the second inequality follows from the fact that $\ln(a + b) \le \ln(a) + b/a$. 
Combining everything yields
\(
\kl{D_{1}}{D_{2}} 
&\le \ln \Lambda_{\nu,k} + \frac{1}{2}\ln\frac{|\Sigma_{2}|}{|\Sigma_{1}|} - \frac{k}{2} \ln 2 e
  +  \frac{\nu + k}{2\nu}\tr(\Sigma_{2}^{-1}\Sigma_{1}) \\
&\phantom{=~} 
 + \frac{\nu + k}{2} \ln\left(1 + \frac{1}{\nu}(\mu_{1} - \mu_{2})^{\top}\Sigma_{2}^{-1}(\mu_{1} - \mu_{2})\right),
\)
where 
\(
\Lambda_{\nu,k} = \left\{ 
  \begin{array}{l l}
    \frac{\nu^{k/2}}{(\frac{\nu + k}{2})^{\underline{k/2}}} & \quad \text{if $k$ is even}\\
    \frac{(\nu + 1)^{1/2}\nu^{(k-1)/2}}{(\frac{\nu}{2})^{1/2}(\frac{\nu + k}{2})^{\underline{(k-1)/2}}} & \quad \text{if $k$ is odd.}
  \end{array} \right.
\)

\subsection{Gaussian and Laplace}
\label{app:gaussian-laplace-kl}

Let $D_{1} = \distNorm(\mu, \sigma^{2})$ and
$D_{2} = \distLaplace(\beta)$.
Then
\(
\kl{D_{1}}{D_{2}} 
&= \ln(2\beta) + \frac{1}{\beta}\EE_{D_{1}}[|x|] - \frac{1}{2}\ln(2\pi e \sigma^{2}) \\
&= \ln(2\beta) + \frac{1}{2\beta}\left[\mu\mathrm{Erf}\left(\frac{\mu}{\sqrt{2}\sigma}\right) + \frac{2\sqrt{2}\sigma}{\sqrt{\pi}} \exp\left\{-\frac{\mu^{2}}{2\sigma^{2}}\right\}\right] - \frac{1}{2}\ln(2\pi e \sigma^{2}) \\
&\le \frac{1}{2}\ln \frac{2\beta^{2}}{\sigma^{2}} + \frac{1}{2\beta}\left[|\mu|\sqrt{1 - \exp\left\{-\frac{2\mu^{2}}{\pi\sigma^{2}}\right\}} + \frac{2\sqrt{2}\sigma}{\sqrt{\pi}} \exp\left\{-\frac{\mu^{2}}{2\sigma^{2}}\right\}\right] - \frac{1}{2}\ln(\pi e).
\)

\section{Proof of \cref{thm:mvt-regret}}

Choose 
$Q_{\btheta^{*},\phi} = \distNorm(\btheta^{*}, \phi^{2}I)$.  
With $P_{0} = \distT_{\nu}(\bzero, \sigma^{2}I)$, we have (\cref{app:gaussian-t-kl})
\(
\kl{Q_{\btheta^{*},\phi}}{P_{0}}
&\le \ln \Lambda_{\nu,n} + \frac{n}{2}\ln\frac{\sigma^{2}}{\phi^{2}} - \frac{n}{2} \ln 2 e
  +  \frac{n(\nu + n)}{2\nu}\frac{\phi^{2}}{\sigma^{2}}
 + \frac{\nu + n}{2} \ln\left(1 + \frac{1}{\nu\sigma^{2}}\|\btheta^{*}\|^{2}\right),
\)
where 
\(
\Lambda_{\nu,n} = \left\{ 
  \begin{array}{l l}
    \frac{\nu^{n/2}}{(\frac{\nu + n}{2})^{\underline{n/2}}} & \quad \text{if $n$ is even}\\
    \frac{(\nu + 1)^{1/2}\nu^{(n-1)/2}}{(\frac{\nu}{2})^{1/2}(\frac{\nu + n}{2})^{\underline{(n-1)/2}}} & \quad \text{if $n$ is odd.}
  \end{array} \right.
\)
Note that if $n$ is even then $\frac{\Lambda_{\nu,n}}{2^{n/2}} \le 1$ 
and if $n$ is odd then $\frac{\Lambda_{\nu,n}}{2^{n/2}} \le \frac{\nu + 1}{\nu}$.
Since $\var_{Q_{\btheta^{*},\phi}}[\theta_{i}] = \phi^{2}$, we have
\(
L_{Bayes}(Z) &
\le \inf_{\btheta^{*}} L_{\btheta^{*}}(Z) 
  + \frac{Tc\phi^{2}}{2} + \frac{n}{2} \ln \frac{\nu + 1}{\nu} 
  + \frac{n}{2}\ln\frac{\sigma^{2}}{\phi^{2}} - \frac{n}{2} 
  +  \frac{n(\nu + n)}{2\nu}\frac{\phi^{2}}{\sigma^{2}} 
  + \frac{\nu + n}{2} \ln\left(1 + \frac{1}{\nu\sigma^{2}}\|\btheta^{*}\|^{2}\right)
\)
Choosing $\phi^{2} = \frac{\nu\sigma^{2}n}{Tc\nu\sigma^{2} + (\nu + n)n}$ 
yields the theorem. 

\section{More on Hierarchical Priors for Sharing Statistical Strength}

\subsection{Multiple Simultaneous Observations}
\label{sec:multi-obs}

The Bayesian learner receives $K$ input-output pairs
$\theset{(\bx_{t}^{(k)}, y_{t}^{(k)})}_{k=1}^{K}$ at each 
time step.
Each output is predicted using a separate weight vector 
$\btheta^{(k)}$, so the $k$-th likelihood is 
$p(y \given \btheta^{(k)} \cdot \bx)$, $k = 1,\dots,K$. 
Write $Z^{(k)} \defined \theset{(\bx_{t}^{(k)}, y_{t}^{(k)})}_{t=1}^{T}$. 
Instead of using independent Gaussian priors 
on $\btheta^{(1)}, \dots,\btheta^{(K)}$,  
place a prior over the means of the $K$ priors. 
For each dimension $j = 1,\dots,n$, let
\[
\mu_{j} \given \sigma_{0}^{2} \dist \distNorm(0, \sigma_{0}^{2})
\]
and
\[
\theta^{(k)}_{j} \given \mu_{j}, \sigma^{2} \dist \distNorm(\mu_{j}, \sigma^{2}), \quad k = 1,\dots,K,
\]
and write
$\btheta_{j}^{(1:K)} \defined (\theta^{(1)}_{j}, \dots, \theta^{(K)}_{j})$.
Integrating out $\mu_{j}$ yields
\[
\btheta_{j}^{(1:K)} \given \sigma_{0}^{2}, \sigma^{2} \dist \distNorm(\bzero, \Sigma), 
\]
where, with $1_{K}$ denoting the $K \times K$ all-ones matrix,
\[
\Sigma &\defined s^{2}\rho 1_{K} + s^{2}(1-\rho)I &
s^{2} &\defined \sigma_{0}^{2} + \sigma^{2} &
\rho &\defined \frac{\sigma_{0}^{2}}{\sigma_{0}^{2} + \sigma^{2}},
\]
The Bayesian learner uses this hierarchical prior to
simultaneously predict $y_{t}^{(1)}, \dots, y_{t}^{(K)}$.
For the following theorem, we must replace \eqref{eq:a2} with
an appropriately modified assumption for the simultaneous
prediction task:
\[
\|\bx_{t}^{(k)}\|_{2} &\le 1 \quad \text{for all $t,k$}. \tag{A2'} \label{eq:a2'}
\]
\bnthm[Hierarchical Gaussian regret, simultaneous observations] \label{thm:hg-regret-simultaneous}
If $\btheta_{j}^{(1:K)} \dist \distNorm(\bzero, \Sigma)$, 
$j = 1,\dots,n$, and \eqref{eq:a2'} holds in lieu of \eqref{eq:a2},
then $\mcR(Z, \btheta^{*})$ is bounded by
\[
\begin{split}
R_{Bayes}^{HG-sim}(Z, \btheta^{*})
&\defined
   \frac{1}{2\gamma^{2}}\textstyle\sum_{k=1}^{K}\|\btheta^{*(k)}\|^{2} 
   + \frac{\sigma_{0}^{2}}{\sigma^{2}\gamma^{2}}\textstyle\sum_{k < \ell}\|\btheta^{*(k)} - \btheta^{*(\ell)}\|^{2} \\
&\phantom{\defined~} + \frac{n}{2} \ln\left(1 + \frac{K\sigma_{0}^{2}}{\sigma^{2}}\right) 
   + \frac{nK}{2} \ln\left(1 - \frac{\sigma_{0}^{2}}{\gamma^{2}} + \frac{Tc\sigma^{2}}{n}\right),
\end{split} \label{eq:l-bayes-hg}
\]
where $\gamma^{2} \defined K\sigma_{0}^{2} + \sigma^{2}$.
\enthm

It is instructive to compare the upper bound given in \eqref{eq:l-bayes-hg}
to $\sum_{k} R_{Bayes}^{G}(Z_{(k)}, \btheta^{*(k)})$ with prior variance 
$s^{2} = \sigma_{0}^{2} + \sigma^{2}$. 
To do so, we find 
$\Delta(\btheta^{*}) \defined \sum_{k} R_{Bayes}^{G}(Z_{(k)}, \btheta^{*(k)}) - R_{Bayes}^{HG}(Z, \btheta^{*})$:
\(
\Delta(\btheta^{*}) 
&= \frac{(K-1)\sigma_{0}^{2}}{2\gamma^{2}s^{2}}\textstyle\sum_{k=1}^{K}\|\btheta^{*(k)}\|^{2} 
- \frac{\sigma_{0}^{2}}{\sigma^{2}\gamma^{2}}\textstyle\sum_{k < \ell}\|\btheta^{*(k)} - \btheta^{*(\ell)}\|^{2} \\
&\phantom{=~} - \frac{nK}{2}\ln\left(\frac{n\frac{s^{2}}{\sigma^{2}}(1 - \frac{\sigma_{0}^{2}}{\gamma^{2}}) + Tcs^{2}}{n + Tcs^{2}}\right)
 - \frac{n}{2} \ln\left(\left[1 + \frac{K\sigma_{0}^{2}}{\sigma^{2}}\right]\frac{\sigma^{2K}}{s^{2K}}\right)
\)
For example, setting $\sigma_{0} = \sigma$, so the correlation 
$\rho$ is $\half$, and $K=2$, 
we find that if 
\(
4\|\btheta^{*(1)} - \btheta^{*(2)}\|^{2} + 6s^{2}n \ln\left(\frac{\frac{4}{3}n + Tcs^{2}}{n + Tcs^{2}}\right)
&\le \|\btheta^{*(1)}\|^{2} + \|\btheta^{*(2)}\|^{2} + 0.863 s^{2}n,
\)
then the hierarchical model has a smaller regret 
bound than the non-hierarchical model.\footnote{
For clarity, we have replaced $3 \ln (4/3)$ with 
the bound $0.863$.}
As long as $Tcs^{2} > 2n$, the condition becomes
$
4\|\btheta^{*(1)} - \btheta^{*(2)}\|^{2} 
\le \|\btheta^{*(1)}\|^{2} + \|\btheta^{*(2)}\|^{2} + C s^{2}n
$
for some $0 < C < 0.863$. 
In this case there are two important observations 
about the benefits of the hierarchical model. 
First, noting that the expected magnitude of 
$\|\btheta^{*(1)}\|^{2}$ and $\|\btheta^{*(2)}\|^{2}$ is $\sigma^{2}n$,
as long as  $\|\btheta^{*(1)}\|^{2}$ and $\|\btheta^{*(2)}\|^{2}$ are
only a constant fraction $C/4$ of their expected magnitudes,
the hierarchical model will always have smaller regret bound. 
Second, even if the previous condition does not hold,
the difference in $\|\btheta^{*(1)} - \btheta^{*(2)}\|$ must be 
significantly larger than the expected magnitudes of
$\|\btheta^{*(1)}\|^{2}$ and $\|\btheta^{*(2)}\|^{2}$ for the hierarchical 
model to have a larger regret bound than the non-hierarchical model. 
Thus, the use of the hierarchical model has potentially
significantly reduced regret compared to the non-hierarchical model.

\subsection{Two-level Prior}
\label{sec:two-level-prior}

In this section we derive bounds for the two-level prior
in the case of sequential observations. 
Recall that the prior is
\[
\bbeta &\dist \distNorm(0,\sigma_{0}^{2}I)  \\
\bmu^{(s)} &\dist \distNorm(\bbeta,\sigma_{1}^{2}I) & s &= 1,\dots,S \\
\btheta^{(k)} &\dist \distNorm(\bmu^{(s_{k})},\sigma_{2}^{2}I) & k &= 1,\dots,K.
\]
Integrating out $\bbeta$, we immediately obtain:
\[
\bmu^{(1:S)}_{i} &\dist \distNorm(\bzero, \Sigma_{\mu}),
\]
where $\Sigma_{\mu} \defined \sigma_{0}^{2}1_{S} + \sigma_{1}^{2}I$.
Writing $\bmu_{i} = \bmu^{(1:S)}_{i}$ and
$\btheta_{i} = \btheta_{i}^{(1:K)}$, we have
\[
\bmat \bmu_{i} \\ \btheta_{i} \emat
&\dist \distNorm\left(\bzero, \Sigma\right),
& \Sigma &\defined 
\bmat \Sigma_{\mu} & \Sigma_{\mu\theta} \\
\Sigma_{\mu\theta}^{\top} & \Sigma_{\theta} \emat.
\]
Hence, 
\[
\btheta_{i} \given \bmu_{i} 
&\dist \distNorm(\Sigma_{\mu\theta}^{\top}\Sigma_{\mu}^{-1}\bmu_{i},
				 \Sigma_{\theta} - \Sigma_{\mu\theta}^{\top}\Sigma_{\mu}^{-1}\Sigma_{\mu\theta}).
\]
Define the matrix $P$ such that $P_{ks} = \ind\{s = s_{k}\}$. 
We therefore have 
$\Sigma_{\mu\theta}^{\top}\Sigma_{\mu}^{-1}\bmu_{i} = P\bmu_{i}$
and hence $\Sigma_{\mu\theta}^{\top} = P\Sigma_{\mu}$,
and furthermore
$\Sigma_{\theta} - \Sigma_{\mu\theta}^{\top}\Sigma_{\mu}^{-1}\Sigma_{\mu\theta}
= \sigma_{2}^{2}I$
and hence $\Sigma_{\theta} = \sigma_{2}^{2}I + P\Sigma_{\mu}P^{\top}$.

Hence, the prior on $\btheta_{i}$ is $P_{0} = \distNorm(\bzero, \Sigma_{\theta})$. 
Choose $Q_{\btheta_{i}^{*},\bphi} = \distNorm(\btheta_{i}^{*}, \diag\bphi)$,
yielding
\[
\kl{Q_{\btheta_{i}^{*},\bphi}}{P_{0}}
&= \frac{1}{2}\left\{\ln \frac{|\Sigma_{\theta}|}{\prod_{k}\phi_{k}^{2}}
- k - \tr(\Sigma_{\theta}^{-1})\sum_{k}\phi_{k}^{2} + (\btheta^{*}_{i})^{\top}\Sigma_{\theta}^{-1}\btheta^{*}_{i}\right\}.
\]
Straightforward calculations show that the regret is bounded by
\[
\sum_{i=1}^{n}(\btheta^{*}_{i})^{\top}\Sigma_{\theta}^{-1}\btheta^{*}_{i}
+ \sum_{k=1}^{K} \frac{n}{2}\ln\left(2\tr(\Sigma_{\theta}^{-1}) + \frac{cT^{(k)}}{n}\right) + \frac{n}{2}\ln|\Sigma_{\theta}|. 
\]

\subsection{Proof of \cref{thm:hg-regret-simultaneous}}

First take $n=1$,
which will later generalize to arbitrary $n$.
Choose $Q_{\btheta^{*(1:K)},\phi} = \distNorm(\btheta^{*(1:K)}, \phi^{2}I)$ and
note that 
\(
|\Sigma| = \sigma^{2K-2}(K\sigma_{0}^{2} + \sigma^{2}) = \sigma^{2K-2}\gamma^{2}
\quad \text{and} \quad 
\Sigma^{-1} = - \frac{\sigma_{0}^{2}}{\sigma^{2}\gamma^{2}}1_{K} +  \frac{1}{\sigma^{2}}I.
\)
Thus (\cref{app:gaussians-kl})
\(
\kl{Q_{\btheta^{*(1:K)},\phi}}{P_{0}}
&= \frac{1}{2}\left\{ \ln\frac{|\Sigma|}{|\phi^{2}I|} - K + \phi^{2}\tr(\Sigma^{-1})
 + (\btheta^{*(1:K)})^{\top}\Sigma^{-1}\btheta^{*(1:K)} \right\} \\
&= \frac{K}{2}\ln\frac{\sigma^{2}\gamma^{2/K}}{\phi^{2}\sigma^{2/K}} - \frac{K}{2} 
 + \frac{K(\gamma^{2} - \sigma_{0}^{2})}{2\sigma^{2}\gamma^{2}}\phi^{2} \\
&\phantom{=~} + \frac{1}{2\gamma^{2}}\sum_{k=1}^{K}(\theta^{*(k)})^{2}
 + \frac{\sigma_{0}^{2}}{\sigma^{2}\gamma^{2}}\sum_{k < \ell}(\theta^{*(k)} - \theta^{*(\ell)})^{2}.
\)
Moving to the case of general $n$, since 
$\var_{Q_{\btheta^{*},\phi}}[\sum_{k}\theta^{(k)}_{j}] = K\phi^{2}$
for all $j = 1,\dots,n$, applying \cref{thm:meta-theorem} gives
\(
L_{Bayes}(Z) 
&\le  \sum_{k=1}^{K} L_{\btheta^{*(k)}}(Z^{(k)}) + \frac{TKc\phi^{2}}{2}
 + \frac{nK}{2}\ln\frac{\sigma^{2}\gamma^{2/K}}{\phi^{2}\sigma^{2/K}} - \frac{nK}{2}  \\
&\phantom{\le~} \frac{nK(\gamma^{2} - \sigma_{0}^{2})}{2\sigma^{2}\gamma^{2}}\phi^{2}
 + \frac{1}{2\gamma^{2}}\sum_{k=1}^{K}\|\btheta^{*(k)}\|^{2}
 + \frac{\sigma_{0}^{2}}{\sigma^{2}\gamma^{2}}\sum_{k < \ell}\|\btheta^{*(k)} - \btheta^{*(\ell)}\|^{2}.
\)
Choosing 
$\phi^{2} = \frac{n\sigma^{2}\gamma^{2}}{n(\gamma^{2}-\sigma_{0}^{2}) + Tc\sigma^{2}\gamma^{2}}$ 
yields the theorem. 

\subsection{Proof of \cref{thm:hg-regret-sequential}}

The proof is similar to that for \cref{thm:hg-regret-simultaneous}. 
However, use separate variances for each source:
\(
Q_{\btheta^{*(1:K)},\bphi}
= \prod_{k} Q_{\btheta^{*(k)},\phi_{k}} 
= \prod_{k}\distNorm(\theta^{*(k)},\phi_{k}^{2}).
\)
The error term from the Taylor expansion used in 
\cref{thm:meta-theorem} is $\sum_{k} \frac{T^{(k)}c\phi_{k}^{2}}{2}$, 
so
\(
L_{Bayes}(Z) 
&\le  \sum_{k=1}^{K} L_{\btheta^{*(k)}}(Z^{(k)}) + \sum_{k} \frac{T^{(k)}c\phi_{k}^{2}}{2}
 + \frac{n}{2}\ln\frac{\sigma^{2K}\gamma^{2}}{\sigma^{2}\prod_{k}\phi_{k}^{2}} - \frac{nK}{2}  \\
&\phantom{\le~} \frac{n(\gamma^{2} - \sigma_{0}^{2})}{2\sigma^{2}\gamma^{2}}\sum_{k}\phi_{k}^{2}
 + \frac{1}{2\gamma^{2}}\sum_{k=1}^{K}\|\btheta^{*(k)}\|^{2}
 + \frac{\sigma_{0}^{2}}{\sigma^{2}\gamma^{2}}\sum_{k < \ell}\|\btheta^{*(k)} - \btheta^{*(\ell)}\|^{2}.
\)
Choosing 
$\phi_{k}^{2} = \frac{n\sigma^{2}\gamma^{2}}{n(\gamma^{2}-\sigma_{0}^{2}) + T^{(k)}c\sigma^{2}\gamma^{2}}$ 
yields the theorem. 

\section{More on Feature Selection}
\subsection{The Bayesian Lasso}
\label{sec:bayesian-lasso}

For Bayesian model average learner we have:
\bnthm[GLM Bayesian lasso regret] \label{thm:lasso-regret}
If $\theta_{i} \dist \distLaplace(\theta_{i}, \beta)$, $i = 1,\dots, n$,
then 
\[
\begin{split}
\mcR(Z, \btheta^{*})
&\le \frac{1}{2\beta}\sum_{i}\min\left\{\sqrt{\frac{2}{\pi\phi^{2}}}(\theta_{i}^{*})^{2},|\theta_{i}^{*}|\right\} \\
&\phantom{\le~} + \frac{n}{2}\ln\left(\frac{2T^{2}c^{2}\beta^{4}}{\left(\sqrt{2n^{2} + Tcn\beta^{2}\pi} - \sqrt{2n^{2}}\right)^{2}}\right). 
\label{eq:lasso-regret-bound}
\end{split}
\]
\enthm

In the regime of $Tc\beta^{2} \ll n$, 
\eqref{eq:lasso-regret-bound} becomes (approximately)
\(
\mcR(Z, \btheta^{*})
&\le \frac{1}{2\beta}\sum_{i}\min\left\{\sqrt{\frac{2}{\pi\phi^{2}}}(\theta_{i}^{*})^{2},|\theta_{i}^{*}|\right\} + C n
\)
for some constant $C$ independent of $\beta$ and $c$. 
Hence, even for sparse $\btheta^{*}$, the regret bound is 
$\Theta(n)$. 
The inequalities used to prove the regret bound are all quite tight,
so we conjecture that, up to constant factors, there is a matching lower bound,
as least in the Gaussian regression case.

\subsection{Proof of \cref{thm:lasso-regret}}

Apply \cref{thm:meta-theorem} with
$Q_{\btheta^{*},\phi} = \distNorm(\btheta^{*}, \phi^{2}I)$.  
Since $p_{0}(\btheta) = \prod_{i}\distLaplace(\theta_{i}, \beta)$, we have
(see \cref{app:gaussian-laplace-kl})
\(
\kl{Q_{\theta^{*},\phi}}{P_{0}} 
&\le \frac{n}{2}\ln \frac{2\beta^{2}}{\phi^{2}} - \frac{n}{2}\ln(\pi e) 
  + \frac{1}{2\beta}\sum_{i}\left[|\theta_{i}^{*}| \sqrt{1 - \exp\left\{-\frac{2(\theta_{i}^{*})^{2}}{\pi\phi^{2}}\right\}} 
  + \frac{2\sqrt{2}\phi}{\sqrt{\pi}} \exp\left\{-\frac{(\theta_{i}^{*})^{2}}{2\phi^{2}}\right\}\right] \\
&\le \frac{n}{2}\ln \frac{2\beta^{2}}{\phi^{2}} - \frac{n}{2}\ln(\pi e) + \frac{\sqrt{2}n\phi}{\sqrt{\pi}\beta} 
+ \frac{1}{2\beta}\sum_{i}\min\left\{\sqrt{\frac{2}{\pi\phi^{2}}}(\theta_{i}^{*})^{2},|\theta_{i}^{*}|\right\}.
\)
Since $\var_{Q_{\btheta^{*},\phi}}[\theta_{i}] = \phi^{2}$, 
\(
L_{Bayes}(Z) &
\le \inf_{\btheta^{*}} L_{\btheta^{*}}(Z) 
  + \frac{Tc\phi^{2}}{2}  - \frac{n}{2}\ln(\pi e) + \frac{\sqrt{2}n\phi}{\sqrt{\pi}\beta} 
  + \frac{n}{2}\ln \frac{2\beta^{2}}{\phi^{2}} 
  + \frac{1}{2\beta}\sum_{i}\min\left\{\sqrt{\frac{2}{\pi\phi^{2}}}(\theta_{i}^{*})^{2},|\theta_{i}^{*}|\right\}.
\)
Choosing 
$\phi^{2} = \frac{\left(\sqrt{2n^{2} + Tcn\beta^{2}\pi} - \sqrt{2n^{2}}\right)^{2}}{T^{2}c^{2}\beta^{2}\pi}$
gives the desired result.

\subsection{Proof of \cref{thm:ss-regret}}

Fix some $\btheta^{*}$. If $\theta_{i}^{*} = 0$, then let 
$Q_{\theta_{i}^{*},\phi^{2}} = \delta_{0}$, so
$\kl{Q_{\theta_{i}^{*},\phi^{2}}}{P_{0}} = \ln \frac{1}{p}$.
If $\theta_{i}^{*} = 0$, then let 
$Q_{\theta_{i}^{*},\phi^{2}} = \distNorm(\theta_{i}^{*},\phi^{2})$,
so 
\(
\kl{Q_{\theta_{i}^{*},\phi^{2}}}{P_{0}} 
= \kl{Q_{\theta_{i}^{*},\phi^{2}}}{\distNorm(0, \sigma^{2})} + \ln \frac{1}{1-p}.
\)
The rest of the proof of \eqref{eq:ss-regret-bound} then closely follows 
earlier ones. 
To obtain \eqref{eq:ss-regret-bound-2}, we observe that if $p = q^{1/n}$, then
\(
m \ln \frac{1}{1-p} = m \ln \frac{1}{1 - q^{1/n}} \le m \ln\frac{n}{1 - q}
\)
and 
\(
(n-m) \ln \frac{1}{p} = \frac{n-m}{n} \ln \frac{1}{q} \le \ln \frac{1}{q}.
\)

\end{document}